\pdfoutput=1

\documentclass[11pt]{article}

\usepackage[final]{acl}

\usepackage{times}
\usepackage{latexsym}

\usepackage[T1]{fontenc}

\usepackage[utf8]{inputenc}

\usepackage{microtype}

\usepackage{inconsolata}

\usepackage{graphicx}
\usepackage{amsfonts}
\usepackage{amsmath}
\usepackage{graphicx,subfigure}
\usepackage{booktabs}
\usepackage{multirow}
\usepackage{url}
\usepackage{tabularx}
\usepackage{makecell}
\usepackage{multirow}
\usepackage{graphicx}
\usepackage{comment}
\usepackage{color}
\usepackage{kotex} 
\usepackage{hyperref}
\usepackage{multicol}
\usepackage{multirow}

\usepackage{stfloats}
\usepackage{placeins}
\usepackage{afterpage}
\usepackage{geometry}
\usepackage{pgfplots} 
\usepackage{stfloats}
\usepackage{placeins}
\usepackage{afterpage}
\usepackage{geometry}
\usepackage{tikz}
\usepackage{subcaption}
\usepackage{lscape}          
\usepackage{transparent}
\usepackage{pifont}

\usepackage{subcaption}
\usepackage{lscape}          

\hypersetup{
    colorlinks=true,
    linkcolor=blue,
    filecolor=magenta,      
}

\newcolumntype{L}[1]{>{\raggedright\arraybackslash}p{#1}}
\usepackage{ragged2e}        
\newcolumntype{J}[1]{>{\justifying\arraybackslash}p{#1}}
\usepackage[skip=8pt]{caption}      
\newcolumntype{M}[1]{%
  >{\justifying\setlength{\parindent}{0pt}\arraybackslash}m{#1}%
}

\definecolor{softpastelgreen}{RGB}{198, 233, 211}
\definecolor{softgreen}{RGB}{144,238,144}
\definecolor{darkgreen}{RGB}{0, 100, 0}
\definecolor{mygreen}{RGB}{34,139,34}

\title{Towards a Holistic and Automated Evaluation Framework \\for Multi-Level Comprehension of LLMs in Book-Length Contexts}

\author{\bf Jiaqi Deng$^{1,}$\thanks{~~Equal Contribution; The order was assigned randomly.},\bf ~ Yuho Lee$^{1,*}$,\bf ~ Nicole Hee-Yeon Kim$^{1}$,\bf ~ Hyangsuk Min$^{1}$,\\{\bf ~ Taewon Yun$^{1}$},{\bf ~ Minjeong Ban$^{1}$},{\bf ~ Kim Yul$^{1}$},{\bf ~ Hwanjun Song$^{1}$\thanks{~~Corresponding Author.}}\\
$^{1}$Korea Advanced Institute of Science and Technology\\
\ \{deng.jiaqi, yuholee, songhwanjun\}@kaist.ac.kr}

\newcommand{\algname}{{\sc HAMLET}}

\definecolor{softpastelgreen}{RGB}{198, 233, 211}  

\newcolumntype{L}[1]{>{\raggedright\let\newline\\\arraybackslash\hspace{0pt}}m{#1}}
\newcolumntype{X}[1]{>{\centering\let\newline\\\arraybackslash\hspace{0pt}}p{#1}}
\newcolumntype{Y}[1]{>{\raggedleft\let\newline\\\arraybackslash\hspace{0pt}}m{#1}}

\begin{document}
 \maketitle

\begin{abstract}
We introduce HAMLET, a holistic and automated framework for evaluating the long-context comprehension of large language models (LLMs). HAMLET structures source texts into a three-level key-fact hierarchy at root-, branch-, and leaf-levels, and employs query-focused summarization to evaluate how well models recall and faithfully represent information at each level. To validate the reliability of our fully automated pipeline, we conduct a systematic human study, showing that our automatic evaluation achieves over 90\% agreement with expert human judgments, while reducing the cost by up to 25 times. HAMLET reveals that LLMs struggle with fine-grained comprehension, especially at the leaf level, and are sensitive to positional effects like the lost-in-the-middle. Analytical queries pose greater challenges than narrative ones, and consistent performance gaps emerge between open-source and proprietary models, as well as across model scales. {Our code and dataset are publicly available at \href{https://github.com/DISL-Lab/HAMLET}{link}.}


\end{abstract}

\section{Introduction}\label{sec:introduction}

\begin{figure*}
    \centering
    \includegraphics[width=1.02\linewidth]{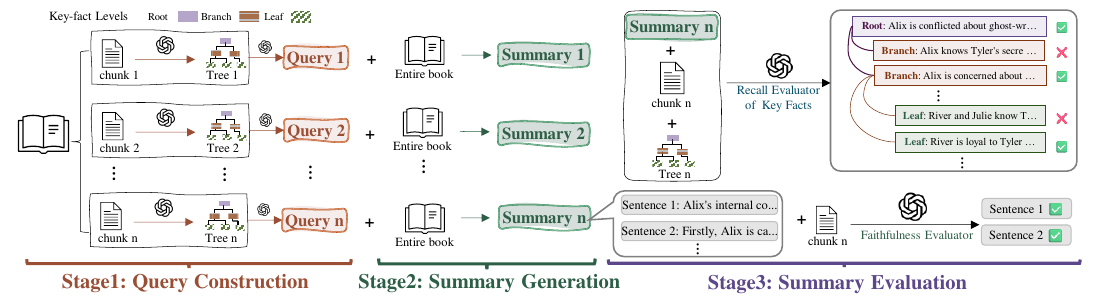}
    \vspace{-0.65cm}
    \caption{Overview of \algname{}: HAMLET constructs queries from a key-fact hierarchy (root, branch, leaf) per chunk, generates summaries using the full book, and evaluates recall and faithfulness using LLM-based evaluators. }
    \label{fig:overview}
\vspace{-0.4cm}
\end{figure*}

As LLMs are increasingly applied to long-form text understanding, recent advances now allow them to process inputs exceeding 100K tokens, enabling comprehension of book-length documents \cite{ding2024longrope, fulazyllm, jin2024llm}. With the growing demand for accurate long-form processing, evaluating the comprehension capabilities of LLMs in book-length texts has become a critical challenge \cite{kryscinski-etal-2022-booksum, zhang-etal-2024-infinitebench}. As with short texts, evaluating book-length comprehension needs evaluating faithfulness, coherence, and others, but poses challenges in holistic evaluation and the high cost of annotation, due to the complexity of extremely long inputs \cite{wu2023less, zhang2024bench, laban-etal-2024-sumhaystack}.

Recent efforts have been made on the evaluation of the LLM's comprehension in a lengthy context, such as {\sc BooookScore} \cite{chang2024booookscore} and {\sc Fables} \cite{kim-etal-2024-fables}. However, existing works remain limited to the \emph{coarse}-grained evaluation of the understanding of LLMs, \emph{i.e.}, short-form generation tasks such as single-turn QA \cite{dong-etal-2024-bamboo,an-etal-2024-leval,kwan-etal-2024-m4le,wang2024novelqa}, or whole-document summarization that only requires a shallow, surface-level understanding of the text \cite{kryscinski-etal-2022-booksum,chang2024booookscore, zhang-etal-2024-infinitebench,kim-etal-2024-fables}. That is, they overlook the LLM's ability to recall information across \emph{varying} levels of detail, an aspect we refer to as \emph{multi-level recall}\footnote{Multi-level recall refers to an LLM's ability to retain and reproduce information at different levels of detail, from high-level summaries to fine-grained facts, in long-form contexts.}. This aspect is especially critical for book-length comprehension, which demands tracking both global themes and specific details and building logical relationships between different levels of information. Without it, LLMs risk generating responses that omit key information or lack coherence \cite{wan2024positional, maharana2024evaluating}.

In this paper, we propose a novel evaluation benchmark framework, \algname{} (\underline{H}olistic and \underline{A}utomated \underline{M}ulti-\underline{L}evel \underline{E}valuation for Long \underline{T}ext) in Figure \ref{fig:overview}, a scalable and automated benchmark framework to evaluate the capabilities of LLMs on book-length contexts across varying levels of detail, in addition to faithfulness. To flexibly probe an LLM's comprehension, we introduce a \emph{key-fact tree}, which is a hierarchical information structure derived from manageable chunks (\emph{i.e.}, 4K-token segments) of long texts. 
Specifically, each tree captures multi-level content abstraction, structuring key-facts into a \emph{root-branch-leaf} hierarchy of themes, supporting ideas, and fine-grained details (see Section \ref{sec:key_fact_tree}). The key-fact tree enables the formulation of detail-aware queries to evaluate an LLM’s ability to extract information across varying levels of abstraction, categorized into two perspectives: \emph{analytical}, which focuses on deeper meaning and thematic interpretation, and \emph{narrative}, which emphasizes story progression and key events. 

Based on these queries, we adopt \emph{query-focused} summarization \cite{xu2020coarse, liu2024querysum}  as the core task, where the LLM receives the entire book as input and generates a summary in response to each query. This setup aligns with the open-ended nature of long-form generation and facilitates fine-grained evaluation of LLMs’ recall and factuality across different levels of abstraction.
Particularly, by anchoring each query to a specific chunk at a distinct location in the book, \algname{} reveals positional challenges faced by LLMs, including insights into how the lost-in-the-middle effect \cite{liu2024lost} manifests differently across levels of information abstraction.

To enable full automation of \algname{}, we conduct a systematic human study to assess the reliability of LLM-based evaluation across its three key components that typically require human verification (see Section~\ref{sec:expert_eval}): {"key-fact tree construction}," ensuring faithfulness, objectivity, and significance of the key-fact tree; {"query validation}," confirming that each query is a self-contained question grounded by the matched key-fact tree; and {"summary evaluation}," assessing summary quality via key-fact alignment and fact-verification to evaluate LLMs’ recall and faithfulness. Our automated pipeline outperforms crowd workers, achieves over 90\% agreement with expert annotations, and operates at up to 25$\times$ lower cost than the off-the-shelf evaluator {\sc FineSurE} \cite{song2024finesure}, showing its reliability and efficiency.

This is notable, as assessing fine-grained comprehension across varying levels of abstraction in book-length contexts is challenging even for humans. Our automation enables scalable benchmarking and allows for effortless extension to new domains with minimal human intervention.

Our main contributions are: (1) we introduce the key-fact tree, a hierarchical abstraction of long texts that enables detail-aware evaluation of LLMs’ comprehension; (2) we adopt query-focused summarization, a natural fit for open-ended evaluation, to probe LLMs’ performance on both recall and faithfulness; (3) we present an automated evaluation pipeline, achieving over 90\% agreement with expert annotations; and (4) we benchmark long-context comprehension across root-branch-leaf abstraction levels, comparing eight LLMs along three axes: open-source vs. proprietary; small vs. large; and non-reasoning vs. reasoning models.

\section{Related Work}\label{sec:relatedworks}


\begin{table*}[t]
\scriptsize
\renewcommand{\arraystretch}{0.4}
\setlength{\tabcolsep}{5.9pt}
\begin{center}
\begin{tabular}{lcclc}
\toprule
\multicolumn{1}{c}{Benchmark} &
  \multicolumn{1}{c}{\makecell{Document\\Authenticity}} &
  \multicolumn{1}{c}{\makecell{Evaluation\\Dimensions}} &
  \multicolumn{1}{c}{\makecell{Evaluated Abstraction\\Level (Task)}} &
  \multicolumn{1}{c}{\makecell{Requires\\Human Annotation}} \\
\midrule
BooookScore~\cite{chang2024booookscore} &
  \textcolor{mygreen}{\ding{51}} Real &
  Coherence &
  ~~~~~High-level (Summary) &
  \textcolor{red}{\ding{55}} Required \\
NoCha~\cite{karpinska-etal-2024-onethousand_nocha} &
  \textcolor{mygreen}{\ding{51}} Real &
  Faithfulness &
  ~~~~~Low-level (Claim Discrimination)  &
  \textcolor{red}{\ding{55}} Required \\
NovelQA~\cite{wang2024novelqa} &
  \textcolor{mygreen}{\ding{51}} Real &
  Faithfulness &
  ~~~~~High-level (Simple QA) &
  \textcolor{red}{\ding{55}} Required \\
FABLES~\cite{kim-etal-2024-fables} &
  \textcolor{mygreen}{\ding{51}} Real &
  Faithfulness &
  ~~~~~High-level (Summary)  &
  \textcolor{red}{\ding{55}} Required \\
$\infty$Bench~\cite{zhang-etal-2024-infinitebench} &
  \textcolor{mygreen}{\ding{51}} Real &
  Faithfulness &
  ~~~~~High-level (Simple QA \& Summary) &
  \textcolor{mygreen}{\ding{51}} Not Required \\
SummHay~\cite{laban2024summary} &
  \textcolor{red}{\ding{55}} Synthetic &
  Faithfulness, Attribution &
  ~~~~~Low-level (Query-focused Summ.) &
  \textcolor{red}{\ding{55}} Required \\
MedOdyssey~\cite{fan2025medodyssey} &
  \textcolor{mygreen}{\ding{51}} Real &
  Recall, Faithfulness &
  ~~~~~Low-level (Multi-choice QA) &
  \textcolor{red}{\ding{55}} Required \\
\midrule
\textbf{HAMLET (Ours)} &
  \textcolor{mygreen}{\ding{51}} \textbf{Real} &
  \textcolor{mygreen}{\ding{51}} \textbf{Multi-level Recall, Faithfulness} &
  \textcolor{mygreen}{\ding{51}} \textbf{High$\rightarrow$Low (Query-focus Summ.)} &
  \textcolor{mygreen}{\ding{51}} \textbf{Not Required} \\
\bottomrule
\end{tabular}
\vspace*{-0.25cm}
\caption{Comparison of \algname{} with seven recent benchmark frameworks evaluating LLMs’ comprehension of book-length contexts with respect to four key criteria: (Document Authenticity) use of real vs. synthetic documents; (Evaluation Dimension) targeted aspects of long-form understanding; (Abstraction Level) level of information abstraction assessed; and (Requires Human Annotation) reliance on human annotations.}
\label{tab:benchmarks_comparison}
\end{center}
\vspace*{-0.6cm}
\end{table*}

\paragraph{Long-form Comprehension Benchmarks.}
Existing LLM benchmarks for long-form comprehension have largely focused on assessing information retrieval through simple QA tasks \cite{wang2024novelqa, karpinska-etal-2024-onethousand_nocha, cheng-ping-etal-2024-ruler}, offering a limited view of an LLM's deeper understanding. Beyond this, several recent benchmarks have tuned to long-form generation tasks, generating a single, holistic summary of the entire book \cite{xu-etal-2023-lmgqs, kwan-etal-2024-m4le, zhang-etal-2024-infinitebench, an-etal-2024-leval, kim-etal-2024-fables, laban-etal-2024-sumhaystack, fan2025medodyssey}. For instance, {\sc BooookScore} \cite{chang2024booookscore} evaluates the coherence and readability of holistic summaries, while {\sc Fables} focuses on faithfulness by identifying unfaithful statements. However, both overlook the diverse abstraction levels in long texts, offering only shallow evaluations and failing to assess LLMs’ recall, with an emphasis limited to coherence and factuality.
In contrast, {\sc SummHay} \cite{laban2024summary} adopts query-focused summarization, but its evaluation is limited to information retrieval performance, without assessing LLMs’ recall or faithfulness. Its fully synthetic data further limits the ability to assess LLMs' real capabilities.

\smallskip\smallskip\noindent
\textbf{{Automated Evaluation.}} A critical challenge in human-based benchmarking is the high cost of human evaluation \cite{kim-etal-2024-fables, lee2024unisumeval}, making the benchmarking pipeline impractical for scalable and repeated use. To address this, many studies explore using LLMs as automated judges to reduce reliance on costly human annotators. 
Specifically for summary evaluation, traditional rule-based approaches \cite{lin2004rouge, zhang2019bertscore} and learning-based metrics \cite{zhong2022towards, achiam2023gpt} have shown low agreement with human judgments and are often confined to specific evaluation dimensions, such as faithfulness or completeness. In contrast, LLM-based metrics have demonstrated much stronger alignment with human assessments and can be easily extended to other dimensions through prompt tuning \cite{liu2023geval, wang2023chatgpt, tang2024minicheck, van2024field, fu2024gptscore}. Among these, {\sc FineSurE} \cite{song2024finesure} enables fine-grained, multi-dimensional evaluation with interpretable scores, such as factual error rates and missing content proportions.

Yet, automatic evaluation of book-length inputs remains challenging, as it requires tracking and understanding complex narratives over long contexts. Consequently, recent works, including {\sc Fables} \cite{kim-etal-2024-fables} and {\sc NovelQA} \cite{wang2024novelqa}, rely mostly on human annotators who have read the book; however, their memory-based judgments are limited to surface-level verification, making fine-grained evaluation infeasible.

\section{HAMLET Framework}\label{sec:pipeline}

We first collect 16 novels,\footnotemark[2] which are recently published books with an average length of 101K tokens. These narrative-rich documents provide a challenging yet realistic testbed for evaluating long-form comprehension. Based on this, we construct our benchmark framework in three stages: (Stage 1) query construction, (Stage 2) summary generation, and (Stage 3) automated summary evaluation. In Section~\ref{sec:humanval}, we validate the reliability of automated components through expert human evaluation.


In contrast to recent benchmarks, \algname{} is the first automated framework to evaluate LLMs’ long-context comprehension on multi-level recall and faithfulness, as summarized in Table \ref{tab:benchmarks_comparison}.

\subsection{Query Construction (Stage 1)} 

To evaluate the comprehension of long context for LLMs, we assess their response accuracy to queries grounded in different parts of a long document, focusing on recall and faithfulness. We construct such queries through three steps: text chunking, key-fact tree construction, and query formulation. These steps ensure that each query is anchored to specific locations within the long text and to different levels of information abstraction.

\footnotetext[2]{The selected books were mostly published after January 2025, ensuring they were not included in the training data of the evaluated LLMs (refer to the list of books in Table~\ref{tab:book_info}). The book list is easily configurable, and our automated framework generalizes seamlessly to new long-form inputs.}

\subsubsection{Text Chunking} 

Given the extreme length of book-scale inputs, generating targeted queries from the full text is impractical for both humans and LLMs. We therefore segment each book into sequential, non-overlapping 4K-token chunks, which serve as localized anchors for key-fact tree construction and query generation. 
{A detailed justification and additional experiments on chunk size are provided in Appendix \ref{appendix:chunk_size}.}

\subsubsection{Key-fact Tree Construction} \label{sec:key_fact_tree}

\begin{figure}
    \centering
    \includegraphics[width=1\linewidth]{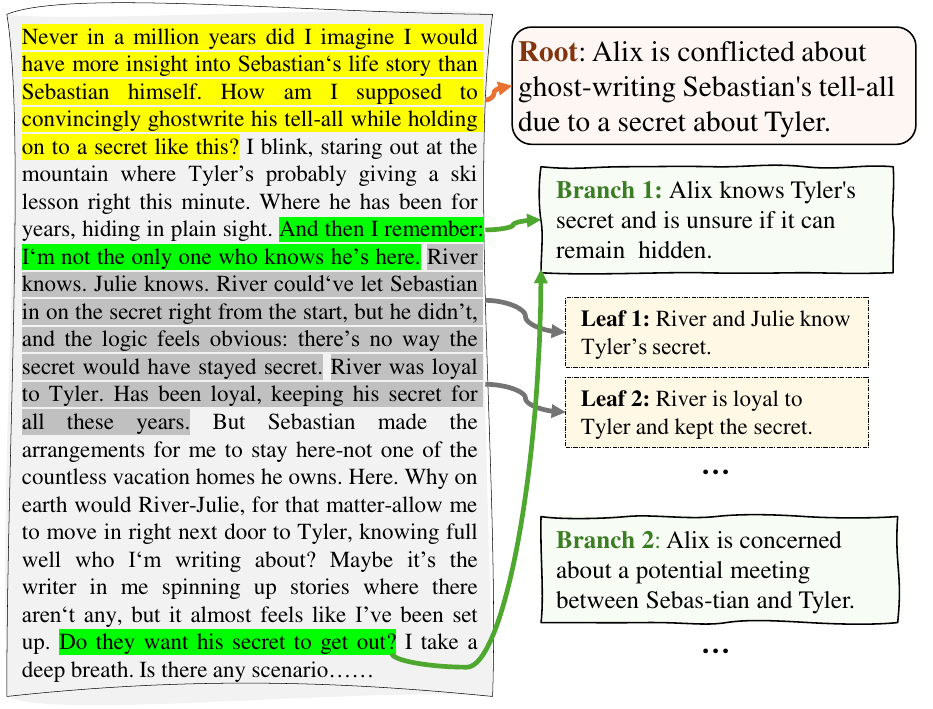}
    \vspace{-0.8cm}
    \caption{Example of an analytical key-fact tree extracted from a single chunk using GPT-4o.}
    \label{fig:key_fact_tree_examples}
\vspace{-0.5cm}
\end{figure}

Next, we extract key information from each chunk at varying levels of detail. Specifically, we organize this information into a hierarchical structure called a \emph{key-fact tree} (refer to Table \ref{tab:keyfact-query-definition} for a detailed definition of each level), which compresses all key information from a book chunk into three semantic levels, as exemplified in Figure \ref{fig:key_fact_tree_examples}. \emph{Roots} capture the chunk’s central theme; \emph{branches} represent conceptual subtopics or supporting narrative arcs; and \emph{leaves} enumerate specific facts or fine-grained evidence associated with each branch. This structure provides a principled way to assess LLM comprehension across abstraction levels.

Importantly, the key-fact tree is central to HAMLET: (i) it guides query generation by defining structure, detail level, and logical relations in Section \ref{sec:query_generation}, and (ii) serves as a reference for evaluating summaries through fine-grained, query-aligned recall and faithfulness analysis in Section \ref{sec:summary_evaluation}. 

\smallskip\smallskip\noindent
\textbf{{Analytical vs. Narrative.}}
To further structure the extracted key-fact trees, we classify them into two types: \emph{analytical}, which emphasizes deeper meaning and logical reasoning, and \emph{narrative}, which focuses on temporal progression and causal relationships. This distinction is essential for constructing key-fact trees, as it allows a structured representation of both reasoning and narrative elements, \emph{i.e.}, two core aspects of long-form comprehension~\cite{mar2021memory}. See Table~\ref{tab:keyfact-query-definition} for definitions and examples of the two key-fact types.

\smallskip\smallskip\noindent
\textbf{{Generation and Verification.}}
We generate the key-fact tree by prompting GPT-4o with carefully designed prompts, each tailored to either the analytical or narrative perspective, shown in Table~\ref{tab:analytical_tree_generation} and Table~\ref{tab:narrative_tree_generation}. The output is returned in JSON format for ease of parsing and structured processing across the root–branch–leaf hierarchy.

All generated trees undergo an {automated} validation process, as {raw} key-fact trees may include hallucinations, subjective inferences, or trivial details due to reliance on a single model. Therefore, we configure GPT-4o as a specialized validator and implement a three-dimensional filter, covering: \emph{faithfulness}, key-facts must be fully supported by the source chunk; \emph{objectivity}, key-facts must be free from speculation or value judgments; and \emph{significance}, key-facts must provide essential insight, not minor statements. See Tables~\ref{tab:keyfact_tree_validation_faith}–\ref{tab:keyfact_tree_validation_sig} for the verification prompts. 
After the validation passes, \texttt{PASS}/\texttt{FAIL} judgments are merged. Any key-fact failing a single dimension is removed, and orphaned descendants are recursively pruned. This filtering ensures reliable key-fact trees for query formulation and summary evaluation.

We obtain 22,333 key-facts across 16 novels, spanning entire books and covering all three levels of detail. See Appendix~\ref{appendix: dataset_statistics} for statistics.

\vspace{-0.1cm}
\subsubsection{Query Formulation}\label{sec:query_generation}

To assess LLMs' long-context comprehension, we generate queries from previously extracted and verified key-fact trees. Each query is produced by GPT-4o, which takes a 4K-token text chunk and its validated key-fact tree as input and is prompted to generate a single, self-contained query. We use distinct prompts for the two types of key-fact trees, {analytical} or {narrative}, as shown in Tables~\ref{tab:analytical_query_generation} and~\ref{tab:narrative_query_generation}. As key-fact trees span the entire book, queries are generated throughout the text, ensuring broad coverage. The formulated queries provide a robust benchmark for evaluating LLMs’ comprehension of both analytical and narrative aspects in long documents.

As a result, we curate 814 queries (407 for each of the two query types) across 16 novels. The example of queries can be found in Table~\ref{tab:keyfact-query-definition}. 

\vspace{-0.1cm}
\subsection{Summary Generation (Stage 2)} \label{sec:summary_generation}

Based on the queries, we adopt \emph{query-focused} summarization~\cite{xu2020coarse, liu2024querysum} as our task, where the LLM is given the full book (up to 114K tokens) as input and generates a summary in response to each query. This task evaluates the LLM’s ability to extract all relevant information (recall) and generate factually accurate responses (faithfulness) to queries over long documents.

To perform this task, we evaluate eight high-performing LLMs, including GPT-4o (Base and Mini), Claude-3.5 (Sonnet and Haiku), Llama-3.1-Instruct (405B and 8B), and Qwen-2.5-32B (Instruct and R1-distill). They are assessed along three axes: proprietary vs. open-source, large vs. small, and general vs. reasoning-optimized. The model specification is provided in Table~\ref{tab:model_overview}.
%

We obtain 4,884 summaries from the entire 814 queries using six summarizers (excluding Qwen), and 632 summaries from a subset of 316 queries using two Qwen models, due to the high inference cost of reasoning models. All summaries are subject to automatic evaluation in our benchmark.

\subsection{Summary Evaluation (Stage 3)}\label{sec:summary_evaluation}

\subsubsection{Evaluation Dimension}

We benchmark LLMs' recall and faithfulness on extremely long texts through their query-based summaries. \emph{Multi-level recall} measures how well the LLM retrieves relevant information across varying levels of detail, while \emph{multi-level faithfulness} assesses the accuracy of that content without hallucination at each level. High scores in both indicate strong long-context comprehension, reflecting the LLM’s ability to faithfully capture information from high-level concepts to fine-grained details.

\smallskip\smallskip\noindent\textbf{{Multi-level Recall.}}
This metric quantifies how well an LLM-generated summary captures the key-facts needed to answer a query, evaluated at root, branch, and leaf levels of the reference key-fact tree. Let $K_{\text{level}}$ denote the set of key-facts at a specific level (level $\in$ \{root, branch, leaf\}) from the reference tree. Then, for each level, recall is defined as:
\begin{equation}
\text{Recall}_{\text{level}} \!=\! |\{k |k \!\in\! S \wedge k \!\in\! K_{\text{level}}\}| / |K_{\text{level}}|,
\label{eq:multi_recall}
\end{equation}
where $k$ is a content unit from the summary $S$.\footnotemark[3]

\footnotetext[3]{Following \citet{song2024finesure}, we treat each summary sentence as a semantic content unit for simplicity.}

\smallskip\smallskip\noindent\textbf{{Multi-level Faithfulness.}}
We first classify each content unit $k$ in the summary $S$ by whether it matches a key-fact in the reference tree. We then label each $k$ as faithful or hallucinated across four levels (level $\in$ \{root, branch, leaf, none\}), where "none" indicates the content unit not aligned with any key-fact in the reference tree. Let $S_{\text{level}}$ denote the set of content units in the summary assigned to a specific level. Then, faithfulness is defined as:
\begin{equation}
\text{Faithfulness}_{\text{level}} \!=\! |S_{\text{level}}^{*}| / |S_{\text{level}}|,
\label{eq:multi_faith}
\end{equation}
where $S_{\text{level}}^{*} \subseteq S_{\text{level}}$ is the subset consisting only of faithful content units at that level.

Note that by ignoring level distinctions and aggregating across all content units, we can easily compute an overall summary-level score.

\subsubsection{Automatic Evaluation}

Although methods like {\sc FineSurE}~\cite{song2024finesure} have improved automated evaluation for summarization, evaluating models on inputs over 100K tokens remains an open challenge. Most long-document benchmarks still depend on human evaluation~\cite{kim-etal-2024-fables, wang2024novelqa}, yet even annotators who have read the book struggle to consistently assess fine-grained content.

\algname{} addresses this limitation by eliminating the need to reference the full document during evaluation. Instead, it anchors the evaluation of each query-based summary to a localized chunk and its associated key-fact tree, enabling fine-grained and scalable assessment without full-text access. This approach not only reduces the complexity and cost of human evaluation but also enhances the effectiveness of automated assessment by decomposing inputs of up to 114K tokens into manageable 4K-token chunks. To support this, we adapt {\sc FineSurE} to our chunk-based pipeline for the two evaluation dimensions, recall and faithfulenss, using GPT-4o as the backbone. The detailed adaptation approach can be found in Appendix~\ref{appendix:eval_details}.

We compare our method against human assessments by crowd workers and automated approaches that require full-book access, and find that \algname{} significantly outperforms them, achieving over 90\% agreement with expert judgments while reducing cost by up to 25$\times$ (see Section~\ref{sec:humanval}).

\section{Validation of HAMLET Pipeline}\label{sec:humanval}

We evaluate the reliability of our three automated components, including (i) key-fact tree construction, (ii) query formulation, and (iii) summary evaluation, by comparing their outputs against expert human judgments. For this study, we recruit three graduate students with C2 English proficiency and NLP expertise as examiners. Expert disagreements are resolved through discussion, resulting in a consensus label. Note that the sample size used in this study ensures over 98\% confidence level with ±5\% margin of error. See Appendix \ref{apppendix:annotator_recruitment} for details.

The three rigorous verification below demonstrate that our automated pipeline achieves high reliability close to expert-based evaluation.

\smallskip\smallskip\noindent\textbf{{Key-fact Tree Generation.}} \label{sec:expert_keyfact}Key-facts are finalized through our three-dimensional verification filter in Section \ref{sec:key_fact_tree}. Thus, we assess the correctness of our filter, which assigns \texttt{Pass}/\texttt{Fail} judgments for faithfulness, objectivity, and significance to each key-fact. The three experts verify the correctness of each judgment on 1,185 key-facts randomly sampled from a total of 23,333. Table~\ref{tab:humaneval_keyfact} reports the results of expert inspection of the automated filtering judgments. Overall, the filter exhibits the average of 96.2\% and 96.4\% accuracy for \texttt{PASS} and \texttt{FAIL} judgments, respectively, demonstrating the robustness of our verification process in cultivating high-fidelity key-fact trees.

\smallskip\smallskip\noindent\textbf{{Query Formulation.}}\label{sec:expert_queries} To evaluate the quality of the query generated in Section \ref{sec:query_generation}, we evaluate query quality along two dimensions: \emph{naturalness}, referring to the query’s fluency, grammaticality, and human-likeness; and \emph{relevance}, indicating whether the answer can be derived from the corresponding 4K-token chunk. Thus, we ask the experts to evaluate whether each query is valid or not for the entire 814 queries. The results confirm that the queries are valid, with 100\% for naturalness and 98.2\% for relevance, respectively. This aligns with prior findings that modern AI models excel at naturalness and relevance \cite{liu2023geval}. Therefore, our pipeline reliably produces valid, high-quality queries without manual intervention.

\newcolumntype{L}[1]{>{\raggedright\let\newline\\\arraybackslash\hspace{0pt}}m{#1}}
\newcolumntype{X}[1]{>{\centering\let\newline\\\arraybackslash\hspace{0pt}}p{#1}}

\begin{table}[t!]
\centering
\scriptsize
\begin{tabular}{X{1.1cm} | X{1.2cm} X{1.2cm} X{1.3cm} |X{0.8cm}}
\toprule
Judgement & Faithfulness & Objectivity & Significance & Mean \\
\midrule
{PASS}    & 99.3\%     & 99.1\%     & 90.3\%  &  96.2\%   \\ 
{FAIL}  & 100\%       & 91.6\%      & 97.5\% &  96.4\% \\
\bottomrule
\end{tabular}%
\vspace*{-0.2cm}
\caption{Accuracy~(\%) of PASS/FAIL judgments by the verification filter in key-fact tree construction.}
\vspace*{-0.5cm}
\label{tab:humaneval_keyfact}
\end{table}

\smallskip\smallskip\textbf{{Summary Evaluation.}} \label{sec:expert_eval}In our framework, the key concern is the accuracy of automated summary evaluation compared to expert human assessments, which serves as a measure of its reliability. Hence, we first collect gold labels from three experts for two binary tasks: \emph{key-fact alignment}, whether each key-fact appears in the summary, and \emph{fact-checking}, whether each summary sentence is supported by the source document; these labels are used to compute the multi-level recall (in Eq. \eqref{eq:multi_recall}) and faithfulness (in Eq. \eqref{eq:multi_faith}) of LLM responses to the query, respectively. For each task, we sample 600 instances to annotate them with expert labels. Then, we compute binary accuracy (bACC) between the labels generated by the automated evaluation and the expert-annotated gold labels, as summarized in Table~\ref{tab:humaneval_summary_auto_evaluation}.

As an automated summary evaluation method, \algname{} differs from {\sc FineSurE} \cite{song2024finesure} in two key aspects: (i) it uses only a single chunk of text for faithfulness evaluation, and (ii) it employs a distinct key-fact extraction strategy (key-fact trees) to assess LLM recall on book-length context. Consequently, \algname{} achieves significantly higher bACC in evaluating LLMs’ recall and faithfulness compared to {\sc FineSurE}, based on expert labels. In particular, for faithfulness, \algname{} greatly improves bACC by anchoring the query to a 4K-token chunk, simplifying the task. This also reduces evaluation API cost from \$10.50 to \$0.53, achieving a 25$\times$ cost saving.

We also compare our automated pipeline with a variant that uses \emph{crowd-sourced} workers instead of LLMs. We collect the label by the majority vote among three Amazon Mturk workers\footnotemark[4] for each instance (refer to Appendix~\ref{appendix:crowd_sourced_anno}). The result in Table~\ref{tab:humaneval_summary_auto_evaluation} show that a strong LLM like GPT-4o significantly outperforms crowd-sourced workers in summary evaluation, achieving over 90\% accuracy. Even a weaker model, GPT-4o-Mini, surpasses the performance of crowd-sourced labels on average.

\footnotetext[4]{The inter-annotator agreement (IAA) scores, measured using Gwet’s AC1~\cite{gwet2008computing}, are 0.53 for the key-fact alignment task and 0.47 for the fact-checking task.}

\newcolumntype{L}[1]{>{\raggedright\let\newline\\\arraybackslash\hspace{0pt}}m{#1}}
\newcolumntype{X}[1]{>{\centering\let\newline\\\arraybackslash\hspace{0pt}}p{#1}}

\begin{table}[t]
\begin{center}
\scriptsize
\begin{tabular}{L{2.9cm} |  X{0.95cm} | X{1.05cm} X{1.05cm} }\toprule
~~~~~~~~~~~~Eval. Method  &  \!\!Reference\!\!  & Recall & \!\!\!\!Faithfulness\!\!\!\! \\ \midrule
\!FineSurE\,(GPT-4o)  & Full Text & N/A  & 52.9\% \\ \midrule 
\!HAMLET\,(Crowd-sourced)\!\!  & Chunk &  86.8\% & 67.8\% \\  
\!HAMLET\,(GPT-4o-Mini)  & Chunk & 94.3\%  &  64.3\% \\ 
\!\textbf{HAMLET\,(GPT-4o)}  & \textbf{Chunk} &  \textbf{98.1\%} & \textbf{91.6\%}\\ \bottomrule
\end{tabular}
\end{center}
\vspace*{-0.35cm}
\caption{bACC (\%) of automated evaluation methods for key-fact alignment (Recall) and fact-checking (Faithfulness) against expert-annotated gold labels. {\sc FineSurE} does not support key-fact extraction for query-focused summarization, so the cell is marked ‘N/A’.}
\label{tab:humaneval_summary_auto_evaluation}
\vspace*{-0.5cm}
\end{table}

\vspace{-0.1cm}
\section{LLMs' Long-context Comprehension}
\vspace{-0.15cm}
To benchmark LLMs' long-context comprehension, we evaluate six LLMs, \emph{i.e.}, two open-source and four proprietary, on book summarization. We further extend our benchmarking to how response length influences LLM recall and to examine the effectiveness of reasoning-optimized models.


\begin{figure*}[t]
\begin{center}
\includegraphics[width=1\textwidth]{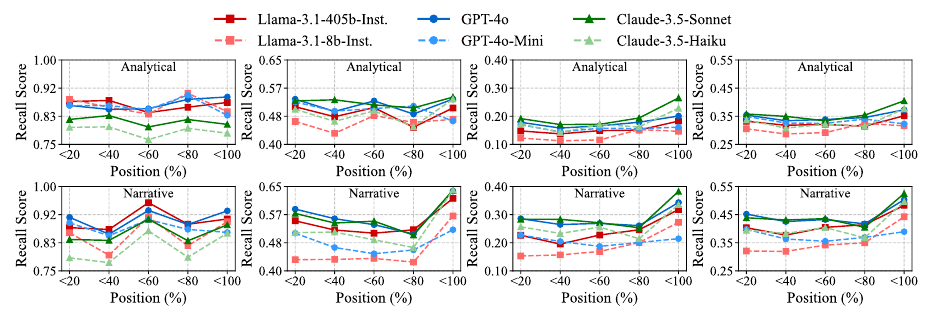} 
\end{center}
\vspace{-0.45cm}
{\small \hspace{1.4cm}(a) Root-level. \hspace{1.8cm} (b) Branch-level. \hspace{1.85cm} (c) Leaf-level. \hspace{2.05cm} (d) All-level.}
\vspace{-0.2cm}
\caption{Recall scores of six LLMs across four key-fact levels, based on query chunk locations for lost-in-the-middle analysis. The first and second rows show recall for analytical and narrative contexts.}
\label{fig:position_wise_tree}
\vspace{-0.15cm}
\end{figure*}

\begin{table*}[t]
\centering
\scriptsize
\setlength{\tabcolsep}{5.7pt}
\begin{tabular}{l|cccc|cccc|cccc|cc
}
\toprule
 & \multicolumn{2}{c}{Llama-3.1-405B} & \multicolumn{2}{c|}{Llama-3.1-8B}  & \multicolumn{2}{c}{GPT-4o} & \multicolumn{2}{c|}{GPT-4o-Mini}  & \multicolumn{2}{c}{Claude-Sonnet} & \multicolumn{2}{c|}{Claude-Haiku}& \multicolumn{2}{c}{Total} \\
\cmidrule(lr){2-3} \cmidrule(lr){4-5} \cmidrule(lr){6-7} \cmidrule(lr){8-9} \cmidrule(lr){10-11} \cmidrule(lr){12-13} \cmidrule(lr){14-15}
 Level & Ana. & Nar. & Ana. & Nar. & Ana. & Nar. & Ana. & Nar. & Ana. & Nar. & Ana. & Nar. & Ana. & Nar. \\
\midrule
Root-level Key-fact	&0.036	&0.079	&0.062	&0.114  &0.037	&0.071	&0.060	&0.048	&0.034	&0.064	&0.038	&0.095	& 0.045 & 0.079\\
Branch-level Key-fact	&0.061	&0.128	&0.052	&0.137 &0.044	&0.128	&0.056	&0.071	&0.032	&0.133	&0.081	&0.171	 & 0.054 & 0.128\\
Leaf-level Key-fact	&0.045	&0.123	&0.040	&0.120 &0.043	&0.082	&0.025	&0.040	&0.095	&0.130	&0.086	&0.121	 & 0.056 & 0.103\\ 
\midrule
All-level Key-fact	&0.037	&0.106	&0.040	&0.123 &0.039	&0.086	&0.026	&0.042	&0.071	&0.120	&0.069	&0.128	 & 0.047 & 0.101 \\

\bottomrule
\end{tabular}
\vspace*{-0.25cm}
\caption{Recall Gap across varying abstraction levels under both analytical (Ana.) and narrative (Nar.) perspectives.}
\label{tab:recall_gap}
\vspace{-0.5cm}
\end{table*}

\vspace{-0.05cm}
\subsection{Main Experiment}
\label{sec:main_exp}
\vspace{-0.05cm}

\subsubsection{Evaluating Multi-level Recall}
\vspace{-0.05cm}
Figure \ref{fig:position_wise_tree} shows LLMs’ recall of key-facts across three abstraction levels, along with overall recall ("all"), computed over five input position splits based on the corresponding queries. Recall is measured by whether each key-fact is included in the generated summary, as mentioned in Eq. \eqref{eq:multi_recall}. 

\smallskip\smallskip\noindent\textbf{{Overview.}} The results show \emph{a consistent decline in recall from high-level gist (root-level in (a)) to fine-grained detail (leaf-level in (c))}. {Considering all LLMs, recall drops from 0.764-0.902 at the root to 0.432-0.540 in the branch, and further to 0.113-0.265 at the leaf for analytical key-facts, while 0.774-0.952 to 0.426-0.640 and further to 0.153-0.383 for narrative ones.} Thus, recent LLMs struggle more to generate responses reflecting detailed, leaf-level information; this limitation is more pronounced with analytical than narrative content.

\smallskip\smallskip\noindent\textbf{{Lost-in-the-Middle.}} LLMs show a pronounced drop in recall for queries linked to chunks appearing between 20\%--80\% of the way through the input, commonly referred to as the lost-in-the-middle effect \cite{liu2024lost}. \emph{Compared to the root- or branch-levels (in (a) and (b)), the drop at the leaf-level (in (c)) is much sharper} with recall up to 0.10 lower than at the document's beginning or end. That is, detailed information is more vulnerable to the lost-in-the-middle effect.

\smallskip\smallskip\noindent\textbf{{Open-source vs. Proprietary LLMs.}} The results demonstrate that {proprietary LLMs generally exhibit higher recall than open-source LLMs across all abstraction levels}, as indicated by the red lines (Llama series) appearing primarily below the others (GPT and Claude series). However, it is noteworthy that the Llama series matches or exceeds proprietary models in root-level recall (in (a)), suggesting that \emph{open-source LLMs are competitive in capturing abstract content but still lag behind in retrieving detailed, fine-grained information.}

\smallskip\smallskip\noindent\textbf{{Large vs. Small LLMs.}} 
We observe substantial and consistent gaps in recall scores between larger and smaller LLMs across all model series (shown by the gap between same-colored solid and dotted lines), highlighting that \emph{larger LLMs are superior for long-context comprehension tasks}. Nevertheless, the two groups show negligible differences in vulnerability to the lost-in-the-middle effect, suggesting that \emph{increasing parameter size alone does not mitigate this issue in recall}.

\smallskip\smallskip\noindent\textbf{{Positional Consistency.}} An important aspect of LLM recall is its positional consistency, indicating robustness to the lost-in-the-middle effect. Table~\ref{tab:recall_gap} reports the recall gap, which is the difference between the highest and lowest recall across five input positions in Figure~\ref{fig:position_wise_tree}. Note that a smaller gap indicates greater positional consistency. 

The GPT series exhibits the best positional consistency. Smaller LLMs tend to be slightly less consistent than their larger counterparts, though the differences are modest. By contrast, \emph{the type of key-fact, analytical or narrative, has a greater impact on positional consistency than model type and size.} Specifically, LLMs exhibit higher recall consistency (\emph{i.e.}, lower recall gaps) on analytical content, likely due to its more explicit and localized structure compared to narrative content.

\subsubsection{Evaluating Multi-level Faithfulness}

Table \ref{tab:levelwise_faithfulness_combined} shows the faithfulness scores of LLM generated summary sentences, labeled as four categories as defined in the Eq. \eqref{eq:multi_faith}. Overall, \emph{hallucinations are frequent across all models}, with faithfulness ranging from {0.151 to 0.674}, much lower than in short-context tasks \cite{lee2024unisumeval, tang2024tofueval}. Moreover, \emph{hallucinations are likely to occur independently of the level of abstraction}, as the scores show little variation across  categories in \{Root, Branch, Leaf\}. However for the "None" category, faithfulness scores are notably lower, suggesting that \emph{content unrelated to any key-facts is highly prone to hallucination especially in narrative contexts}, highlighting a unique challenge in LLMs’ long-context comprehension.

From the model perspective (see the "All" row), \emph{GPT-4o exhibits the highest faithfulness scores}, while Llama-3.1-405B surpasses the proprietary Claude. Interestingly, Claude shows little difference between its small and large versions, unlike other models where the gap is substantial. This suggests that \emph{model size alone does not reliably predict factual alignment}; architecture and alignment objectives may play a greater role in faithfulness. 



\begin{table*}[t]
\scriptsize
\setlength{\tabcolsep}{6.75pt}
\begin{tabular}{l|cccc|cccc|cccc|cc
}
\toprule
Sentence & \multicolumn{2}{c}{Llama-3.1-405B} & \multicolumn{2}{c|}{Llama-3.1-8B} & \multicolumn{2}{c}{GPT-4o} & \multicolumn{2}{c|}{GPT-4o-Mini}  & \multicolumn{2}{c}{Claude-Sonnet} & \multicolumn{2}{c|}{Claude-Haiku}  & \multicolumn{2}{c}{Total} \\
\cmidrule(lr){2-3} \cmidrule(lr){4-5} \cmidrule(lr){6-7} \cmidrule(lr){8-9} \cmidrule(lr){10-11} \cmidrule(lr){12-13} \cmidrule(lr){14-15}
Category & Ana. & Nar. & Ana. & Nar. & Ana. & Nar. & Ana. & Nar. & Ana. & Nar. & Ana. & Nar. & Ana. & Nar. \\
\midrule
Root  & 0.615 & 0.540 & 0.558 & 0.437 & 0.643 & 0.547 & 0.558 & 0.408  & 0.533 & 0.507 & 0.540 & 0.568  & 0.575 & 0.501 \\
Branch  & 0.584 & 0.599 & 0.557 & 0.465 & 0.633 & 0.631 & 0.546 & 0.456  & 0.518 & 0.560 & 0.586 & 0.639 & 0.571 & 0.558 \\
Leaf  & 0.616 & 0.648 & 0.571 & 0.515 & 0.655 & 0.664 & 0.547 & 0.482  & 0.539 & 0.596 & 0.611 & 0.674  & 0.590 & 0.597 \\
None  & 0.497 & 0.243 & 0.404 & 0.151 & 0.507 & 0.378 & 0.464 & 0.281  & 0.481 & 0.317 & 0.382 & 0.272  & 0.456 & 0.274 \\
\midrule
All   & 0.602 & 0.594 & 0.560 & 0.468 & 0.642 & 0.614 & 0.551 & 0.446 & 0.529 & 0.553 & 0.578 & 0.627 & 0.577 & 0.550 \\
\bottomrule
\end{tabular}
\vspace*{-0.2cm}
\caption{Faithfulness scores of summary sentences labeled under analytical and narrative perspectives.}
\label{tab:levelwise_faithfulness_combined}
\vspace{-0.25cm}
\end{table*}

\begin{table*}[t]
    \centering
    \scriptsize
    \setlength{\tabcolsep}{4.9pt}
    \begin{tabular}{l|ccc|c|cccc|c}
    \toprule
       ~~~~~~~~~~~~~~~~~~~~~~~~~~~~~~ Evaluation Dimension &\multicolumn{4}{c|}{Multi-level Recall} & \multicolumn{5}{c}{Multi-level Faithfulness} \\
        \cmidrule(lr){2-5} \cmidrule(lr) {6-10}
         ~~~~~~~~~~~~~~~~~~~~~~~~~~~~~~~~~~~~~~Abstraction Level & Root-level & Branch-level & Leaf-level & All & Root-level & Branch-level & Leaf-level & None  & All \\
         \midrule
         Difference ( \{R1-Distil-Qwen\} - \{Qwen\}) & -0.049& -0.089 & -0.062 & -0.069& 0.194& 0.137 & -0.001 &  0.253 & 0.128\\
         \bottomrule
    \end{tabular}
    \vspace*{-0.25cm}
    \caption{Recall and faithfulness score differences between Qwen2.5-32B-Instruct and DeepSeek-R1-Distil-Qwen-32B across varying abstraction levels. Negative values indicate the non-reasoning model performs better.}
    \label{tab:diff_deepseek_qwen}
    \vspace*{-0.45cm}
\end{table*}

\vspace*{-0.05cm}
\subsection{Additional Experiment}
\label{sec:sub_exp}
\vspace*{-0.15cm}
    
\smallskip\smallskip\noindent\textbf{{General vs. Reasoning Model.}}
As reasoning models show significant performance gains on complex tasks, we examine whether similar benefits hold for long-context tasks. 
Table \ref{tab:diff_deepseek_qwen} shows recall and faithfulness differences between a reasoning model (R1-Distil-Qwen) and its non-reasoning version (Qwen), where the scores are aggregated for each abstraction levels. Contrary to our expectations, recall declines consistently, suggesting \emph{reasoning-optimized training may hinder long-context comprehension} by prioritizing inference over extraction. But, this trade-off improves {faithfulness}, leading to more factually accurate responses.

\smallskip\noindent\textbf{{Impact of Response Length on Recall.}} A potential factor affecting LLM recall is the length of the generated response, since a longer summary leads to greater recall. Hence, we compute the Pearson correlation between summary length and recall scores at each abstraction level, combining outputs from all LLMs. The results demonstrate that \emph{response length has minimal impact on recall, with correlation coefficients close to zero across all levels.} Specifically, the correlation is 0.02, -0.02, -0.09, and -0.07 at root-, branch-, leaf-, and all-levels.


\begin{table}[t!]
\scriptsize
\setlength{\tabcolsep}{9.3pt}
\begin{tabular}{l|cc|c}
\toprule
Model & Recall & Faithfulness & Mean\\
\midrule
Llama-3.1-405B-Instruct  & 0.372 & 0.473 & 0.423\\
Llama-3.1-8B-Instruct    & 0.330 & 0.365 &  0.348\\
\midrule
GPT-4o                   & 0.398 & 0.528 & 0.463\\
GPT-4o-Mini              & 0.354 & 0.420 & 0.387\\
Claude-3.5-Sonnet        & 0.403 & 0.457 & 0.430\\
Claude-3.5-Haiku         & 0.370 & 0.478 & 0.424\\       
\bottomrule
\end{tabular}
\vspace*{-0.25cm}
\caption{Summary-level recall and faithfulness of two open-source LLMs and four proprietary LLMs.}
\label{tab:comp_faith_avg}
\vspace*{-0.5cm}
\end{table}

\smallskip\noindent\textbf{{Summary-level Evaluation.}}
While Section \ref{sec:main_exp} presents fine-grained benchmarking via key-fact and sentence-level analyses, we also conduct a coarse-grained summary-level evaluation to assess how well LLMs generate query-focused summaries for long-context inputs. Table \ref{tab:comp_faith_avg} compares the summary quality of six LLMs in terms of their summary-level recall and faithfulness scores. 
Proprietary and larger LLMs generate better query-focused summaries from long-context inputs. In detail, GPT-4o achieves the highest mean score of 0.463, while the Claude series performs comparably to, or slightly better than, the Llama series. A consistent trend across all three model series is that larger LLMs produce higher-quality summaries than their smaller counterparts.

\vspace*{-0.05cm}
\section{Conclusion}\label{sec:conclusion}
\vspace*{-0.15cm}


We presented a holistic benchmark, automatically evaluating LLMs' long-context comprehension. It is built around a key-fact tree, a root-branch-leaf hierarchy enabling multi-level analysis from analytical and narrative aspects, revealing insights, \emph{e.g.}, the impact of chunk position on LLM comprehension. In particular, its fully automated pipeline outperforms crowd-workers, achieving over 90\% agreement with experts, and reduces cost by 25$\times$.

\section*{Limitations}\label{sec:limitations}

While \algname{} provides significant advancements in evaluating LLMs on book-length texts, several limitations present opportunities for future work. 
First, our benchmark currently focuses on literary novels, and expanding to additional domains such as academic texts, technical documentation, or non-fiction would broaden applicability and test comprehension across different writing styles. However, one practical reason we chose novels as our initial domain is the scarcity of publicly accessible, high-quality documents exceeding 100k tokens in length. Novels offer one of the few consistently high-quality and extensive textual resources freely accessible for research. This choice ensures that our benchmarking pipeline can robustly evaluate LLM comprehension on genuinely long-form content without compromising textual integrity or quality. Such extensive, quality-assured texts are significantly less common or unavailable in many other domains, especially when public accessibility is required for reproducibility and transparency in benchmarking research.
Additionally, established long-context understanding benchmarks such as {\sc BooookScore}~\cite{chang2024booookscore}, {\sc Fables}~\cite{kim-etal-2024-fables}, and {\sc NovelQA}~\cite{wang2024novelqa} have similarly relied primarily on novels. Aligning with these established benchmarks enables clearer comparison and positions our work effectively within the existing research landscape. 
%
Second, the current coverage is limited to English; extending \algname{} to more languages will test multilingual robustness and identify challenges specific to different linguistic structures. 
Third, \algname{} focuses primarily on recall and faithfulness as evaluation dimensions; incorporating safety dimensions, such as bias and toxicity, would provide a more comprehensive assessment of LLM capabilities in long-form text understanding. These limitations outline straightforward directions for strengthening  \algname{} to cover a broader range of evaluation scenarios.

\section*{Future Work}\label{sec:limitations}

{In future work, we plan to extend the \algname{} beyond fictional novels to additional long-form domains such as scientific literature, technical manuals, and long dialogues, enabling broader evaluation of LLM comprehension across diverse writing styles. We will also explore multilingual extensions, leveraging the framework’s language-agnostic design to assess long-context understanding across different linguistic structures. In terms of evaluation dimensions, we aim to expand beyond recall and factual faithfulness to incorporate coherence, reasoning quality, and abstraction ability, with careful design of automated metrics that align with human judgments. 

Furthermore, we plan to investigate weighting schemes for key-fact recall that account for fact significance and hierarchical level, making evaluation more sensitive to the impact of critical information. Finally, we will include qualitative analyses comparing LLM-generated summaries with human-written ones, providing additional insights into model strengths and weaknesses.}
\section*{Ethics Statement}\label{ethics}

Our research placed strong emphasis on transparent communication with all human annotators involved in the evaluation process. We ensured fair compensation practices, providing crowdsourced workers with payments that surpassed U.S. federal minimum wage standards, while our expert evaluators received professional-level compensation (over \$30 hourly) plus additional incentives according to the quality of their work. We maintained strict data privacy protocols throughout the study, carefully anonymizing all personal identifiers in our dataset to protect annotator confidentiality.

\section*{Scientific Artifacts}\label{artifacts}

Our benchmark utilizes 16 commercially published novels with appropriate copyright considerations. For summary generation, we used commercial APIs such as OpenAI and AWS Bedrock. Summary model details are in Table~\ref{tab:model_overview}, providing comprehensive specifications of context window sizes, knowledge cutoffs, and model versions used.
\section*{Acknowledgements}

This work was supported by the IITP grant funded by the Korea government(MSIT) (RS-2025-25410841, Beyond the Turing Test: Human-Level Game-Playing Agents with Generalization and Adaptation) and by the NRF grant funded by Ministry of Science and ICT (RS-2022-NR068758, Industry and Society Demand Oriented Open Human Resource Development).
For GPU infrastructure, our work was supported by the IITP grant funded by MSIT (No. RS-2025-02653113, High-Performance Research AI Computing Infrastructure Support at the 2 PFLOPS Scale).


\clearpage

\appendix
\newpage
\section{Dataset Statistics}\label{appendix: dataset_statistics}

\begin{table*}[ht]
\centering
\resizebox{\textwidth}{!}{%
\begin{tabular}{lccccccccccccc}
\toprule
\makecell[c]{\multirow{2}{*}{Name}} & \multirow{2}{*}{Author} & \multirow{2}{*}{Genre} & \multirow{2}{*}{Token Count} & \multirow{2}{*}{Publication Date} & \multirow{2}{*}{\# Chunks*} & \multicolumn{4}{c}{Analytical} & \multicolumn{4}{c}{Narrative} \\ \cline{7-14} 
 &  &  &  &  &  & Root & Branch & Leaf & Total & Root & Branch & Leaf & Total \\ \hline
Wyoming Burn & Jerry Fedora & Mystery & 74,392 & Jan 2, 2025 & 19 & 56 & 141 & 270 & 467 & 65 & 164 & 302 & 531 \\
No Place Left to Hide & Megan Lally & Thriller & 85,616 & Jan 7, 2025 & 22 & 58 & 169 & 365 & 592 & 68 & 182 & 351 & 601 \\
Lady's Steed & Eve Langlais & \makecell{Romance, \\Fantasy} & 89,884 & Dec 24, 2024 & 23 & 61 & 181 & 384 & 626 & 69 & 207 & 382 & 658 \\
\makecell[l]{The Assassin’s Guide \\to Babysitting} & Natalie C. Parker & Young Adult & 93,337 & Jan 7, 2025 & 24 & 77 & 202 & 387 & 666 & 82 & 225 & 407 & 714 \\
\makecell[l]{A Conventional Boy:\\ A Laundry Files Novel} & Charles Stross & \makecell{Fantasy, \\Horror} & 96,274 & Jan 7, 2025 & 24 & 56 & 181 & 379 & 616 & 74 & 210 & 394 & 678 \\
Lies on the Serpent's Tongue & Kate Pearsall & \makecell{Fantasy,\\ Horror} & 96,311 & Jan 7, 2025 & 24 & 73 & 192 & 377 & 642 & 84 & 209 & 377 & 670 \\
All the Water in the World & Eiren Caffall & Science Fiction & 98,082 & Jan 7, 2025 & 25 & 65 & 195 & 413 & 673 & 97 & 251 & 402 & 750 \\
Holmes is Missing & James Patterson & \makecell{Mystery,\\ Thriller} & 98,497 & Jan 2, 2025 & 25 & 87 & 214 & 382 & 683 & 101 & 246 & 391 & 738 \\
The Lodge & Kayla Olson & Romance & 104,102 & Jan 7, 2025 & 26 & 81 & 205 & 375 & 661 & 98 & 231 & 380 & 709 \\
Switching Graves & Jen Stevens & \makecell{Dark,\\ Gothic} & 107,565 & Jan 3, 2025 & 27 & 85 & 230 & 461 & 776 & 90 & 237 & 417 & 744 \\
So Not My Type & Dana Hawkins & Romance & 107,728 & Dec 12, 2024 & 27 & 66 & 186 & 375 & 627 & 75 & 220 & 384 & 679 \\
Close Your Eyes & Teresa Driscoll & Thriller & 107,933 & Jan 1, 2025 & 27 & 90 & 240 & 489 & 819 & 99 & 283 & 528 & 910 \\
Kingdom of Faewood & Krista Street & Fantasy & 108,780 & Jan 3, 2025 & 28 & 92 & 225 & 447 & 764 & 92 & 220 & 424 & 736 \\
\makecell[l]{Bitter Passage: \\ An Allegheny Beckham Novel} & Colin Mills & Mystery & 111,603 & Jun 22, 2024 & 28 & 72 & 220 & 456 & 748 & 95 & 241 & 460 & 796 \\
The Three Lives of Cate Kay & Kate Fagan & \makecell{Fiction, \\ Mystery} & 113,253 & Jan 7, 2025 & 29 & 105 & 248 & 459 & 812 & 104 & 255 & 424 & 783 \\
Some Other Time & Angela Brown & Fiction & 114,761 & Jan 1, 2025 & 29 & 85 & 223 & 436 & 744 & 99 & 235 & 386 & 720 \\ 
\midrule
\makecell{Average\\(Total)} & - & - & 100,507 & - & \makecell{25\\(407)} & \makecell{76\\(1209)} & \makecell{203\\(3252)} & \makecell{403\\(6445)} & \makecell{682\\(10916)} & \makecell{87\\(1382)} & \makecell{226\\(3616)} & \makecell{401\\(6409)} & \makecell{714\\(11417)} \\ 
\bottomrule
\end{tabular}%
}
\caption{Statistics and details of the novels used in \algname{}. For each chunk, two key-fact trees and two queries are created corresponding to the two summarization perspectives (analytical and narrative).}
\label{tab:book_info}
\end{table*}



Our benchmark consists of 16 novels, each divided into sequential chunks of approximately 4K tokens with preserved sentence boundaries. Table~\ref{tab:book_info} summarizes the basic statistics of \algname{}. For text processing, we use OpenAI's \texttt{tiktoken}\footnote{\url{https://github.com/openai/tiktoken}} tokenization library to compute token counts. Due to copyright restrictions, we release only the generated queries, key-fact trees, model-generated summaries, and evaluation labels, excluding the original book contents.

\section{Choice of Chunk Size}
\label{appendix:chunk_size}
{In this paper, we choose 4K as the chunk size. This size preserves coherence for hierarchical key-facts while remaining short enough for reliable positional evaluation of LLM recall, aligning with recent findings \cite{hescaling}. To validate this choice, we conducted a comparative analysis across four chunk sizes (1K, 2K, 4K, and 8K) using an LLM-as-a-judge framework. The evaluation covered three key dimensions: \emph{validity}, whether the extracted key-facts form a clear multi-level hierarchical structure; \emph{coherence}, the structural integrity and logical flow of the chunk; and \emph{cross-content}, the extent to which the chunk supports reasoning across different sections. Specifically, we first chunked five selected books using three different sizes (1K, 2K, 4K and 8K tokens), and then randomly sampled 100 chunks for each chunk size. Next, we employed two LLMs (GPT-4o and Claude 3.5 Sonnet) as judges to evaluate each chunk across three dimensions, using the standardized prompt shown below. The 1-5 likert-scale scores for each dimension were averaged over the two judges. The detailed prompt can be found in Table~\ref{tab:chunk_size_prompt}.

{Table~\ref{tab:chunk_size_result} shows that scores across all three dimensions improve as the chunk size increases. While performance continues to rise up to 8K, the improvement from 4K to 8K is marginal, indicating that performance largely saturates at 4K. Also, the 4K-token chunk size achieves a high validity score of 4.48 (out of 5.0) for capturing hierarchical key-facts (i.e., root–branch–leaf structure) and a strong coherence score of 4.42 (out of 5.0), suggesting that chunks remain coherent and do not arbitrarily cut across scenes, dialogues, or paragraphs. These findings provide strong empirical support for 4K-token chunks as a meaningful unit of analysis. To sum up, 4K is a more practical choice, as it provides sufficient context to maintain high validity and coherence for hierarchical key-facts, while remaining short enough to enable reliable positional evaluation of LLM recall.}

\begin{table}
    \centering
    \small
    \begin{tabular}{cccc}
    \toprule
        Chunk Size & Vadility & Coherence & Cross-content\\
        \midrule
         1K-token& 2.32 & 3.85 & 3.21\\
         2K-token& 3.50 & 4.15 & 4.20\\
         4K-token& 4.48 & 4.42 & 4.77\\
         8K-token& 4.76 & 4.40 & 4.84\\
         \bottomrule
    \end{tabular}
    \caption{Scores of chunks across different chunk sizes.}
    \label{tab:chunk_size_result}
\end{table}

\section{Key-fact Tree \& Query Type Details} \label{appendix:keyfact_query}

\begin{table*}[ht!]
  \centering
  \scriptsize{}
  \renewcommand{\arraystretch}{0.8}
  \setlength{\tabcolsep}{3pt}
  \begin{tabular*}{\textwidth}{@{\extracolsep{\fill}}
     >{\centering\arraybackslash}m{0.12\textwidth}  
     M{0.4\textwidth}                               
     M{0.4\textwidth}                               
  }
  \toprule
  & \multicolumn{1}{c}{Analytical}
    & \multicolumn{1}{c}{Narrative} \\
  \midrule
  Definition
    & Analytical perspective focuses on thematic elements, implications, character development, and symbolic patterns, examining how these literary components build toward deeper meaning and authorial intent.
    & Narrative perspective emphasizes chronological storytelling, key events, and plot developments, concentrating on how the storyline unfolds and progresses through the text. \\
  \midrule

  Root Definition 
  & A single concise sentence summarizing the overarching purpose, argument, or main analytical insight of the text.
  & A single concise sentence capturing the main idea or overarching sequence of events in the text.
  \\
  \midrule

  Branch Definition 
  & Key supporting ideas, arguments, or elements that develop the overarching purpose or insight, including significant stages, relationships, or turning points in the text. 
  & Key supporting events or developments that progress the narrative logically, including major stages, actions or transitions.
  \\
  \midrule

  Leaf Definition
  & Specific evidence, minor details, or examples that provide additional support or elaboration for each branch. 
  & Specific details, minor events, or pieces of evidence that provide additional clarity or elaboration for each branch. 
  \\
  \midrule
  Query Example
    & How do the social rejection and personal challenges Holmes faces at the bar, his subsequent return home with Callie Brett's assistance, and the professional and personal challenges faced by Poe and Holmes, including the discovery of twins, collectively illustrate the overarching themes of vulnerability, trust, and commitment in the narrative?
    & Provide a detailed summary of the events involving Holmes at the bar, his interaction with Callie Brett, the subsequent health incident, and the developments with Poe and Grey, including their medical appointment and the team's commitment to a crime-writers' convention? \\
  \bottomrule
  \end{tabular*}
  \caption{Definitions of two summarization perspectives and their example queries.}
  \label{tab:keyfact-query-definition}
\end{table*}

Table~\ref{tab:keyfact-query-definition} shows the detailed definitions of each level in our key-fact hierarchy tree and the corresponding queries across analytical and narrative perspectives. Next, the automated key-fact tree validation process filters out about 4\% of root-level key-facts, 13\% of branch-level key-fact, and 30\% of leaf-level key-facts, resulting in an overall pruning rate of 22\%.

\section{Model Details}\label{appendix:model_details}

\begingroup
  \setlength{\tabcolsep}{1pt}%

  \begin{table}[t]
    \centering
    \tiny
     \setlength{\tabcolsep}{1pt}%
      \begin{tabular}{@{}ccccc@{}}
        \toprule
        Model Type &
          Model Name &
          \begin{tabular}[c]{@{}c@{}}Context \\ Length \end{tabular}  &
          \begin{tabular}[c]{@{}c@{}}Knowledge \\ Cutoff \end{tabular} &
          \begin{tabular}[c]{@{}c@{}}Hugging Face Checkpoints \\ \& Official API Version\end{tabular} \\ 
        \midrule
        \multirow{4}{*}{Proprietary} &
          GPT-4o &
          128K &
          Oct, 2023 &
          \begin{tabular}[c]{@{}c@{}}gpt-4o-\\ 2024-08-06\end{tabular} \\
        & GPT-4o-Mini & 128K & Oct, 2023 &
          \begin{tabular}[c]{@{}c@{}}gpt-4o-mini-\\ 2024-07-18\end{tabular} \\
        & Claude-3-5-Sonnet & 200K & Apr, 2024 &
          \begin{tabular}[c]{@{}c@{}}claude-3-5-sonnet-\\ 20240620\end{tabular} \\
        & Claude-3-5-Haiku & 200K & Jul, 2024 &
          \begin{tabular}[c]{@{}c@{}}claude-3-5-haiku-\\ 20241022\end{tabular} \\
        \midrule
        \multirow{4}{*}{Open-Source} &
          LLaMA-3.1-405B & 128K & Dec, 2023 &
          \begin{tabular}[c]{@{}c@{}}meta-llama/\\ Llama-3.1-405B-Instruct\end{tabular} \\
        & LLaMA-3.1-8B & 128K & Dec, 2023 &
          \begin{tabular}[c]{@{}c@{}}meta-llama/\\ Llama-3.1-8B-Instruct\end{tabular} \\
        & Qwen2.5-32B-Instruct & 128K & – &
          \begin{tabular}[c]{@{}c@{}}Qwen/\\ Qwen2.5-32B-Instruct\end{tabular} \\
        & \begin{tabular}[c]{@{}c@{}}DeepSeek-R1-Distill\\ -Qwen-32B \end{tabular} & 128K & – &
          \begin{tabular}[c]{@{}c@{}}deepseek-ai/\\ DeepSeek-R1-Distill-Qwen-32B\end{tabular} \\
        \bottomrule
      \end{tabular}%
    \caption{Overview of model specifications including their types, context window sizes, and knowledge cutoff dates. Note that GPT-4o-Mini is also used as the baseline model for automated summary evaluation (Section~\ref{tab:humaneval_summary_auto_evaluation}).}
    \label{tab:model_overview}
  \end{table}
\endgroup

We compare various proprietary and open-source LLMs, highlighting their context lengths and knowledge cutoffs. The details of these models are presented in Table~\ref{tab:model_overview}. For open-source LLMs, we use instruction-tuned versions. For proprietary models, we use official APIs. To ensure the reproducibility，we use the greedy decoding by setting the temperature parameter to 0. {We spent a total of \$800 on model inference. Due to the long input size, the cost remains significant even before factoring in annotation expenses.}

\section{Evaluation Details}\label{appendix:eval_details} 
Our work builds on \citet{song2024finesure}, adapting their automatic summarization evaluation framework to suit our task. The evaluation consists of two tasks: \textit{Fact verification} and \textit{Key-fact alignment}.
\subsection{Fact Verification}
To evaluate faithfulness of a summary, the summary is split into individual sentences, and the original text from which the query was generated is provided as context. LLM as an evaluator assigns a binary label to each sentence: faithful (1) or unfaithful (0). Additionally, following the taxonomy in Table~\ref{tab:error-categories}, each unfaithful sentence is categorized, with an accompanying explanation. We use GPT-4o for the evaluation. Table~\ref{tab:fact_checking_eval} provides the fact verification prompt.

\begin{table}[t]
\centering
\resizebox{\columnwidth}{!}{%
\begin{tabular}{@{}lll@{}}
\toprule
\makecell[c]{Error Category} & \makecell[c]{Error Type}     & \makecell[c]{Description}                                                                                                                     \\ \midrule
Extrinsic Error &
  \begin{tabular}[c]{@{}l@{}}Out-of-article \\ Error\end{tabular} &
  \begin{tabular}[c]{@{}l@{}}The summary introduces facts, opinions, \\ or information not found in or reasonably \\ inferrable from the text\end{tabular} \\ \midrule
\multirow{3}{*}{Intrinsic Error} &
  Entity Error &
  \begin{tabular}[c]{@{}l@{}}Incorrect reference to key subjects/objects \\ (\emph{e.g.}, wrong names, numbers, pronouns)\end{tabular} \\
               & Relation Error & \begin{tabular}[c]{@{}l@{}}Mistakes in semantic relationships \\ (e.g., incorrect verbs, prepositions, adjectives)\end{tabular} \\
               & Sentence Error & \begin{tabular}[c]{@{}l@{}}Multiple errors causing an entire sentence to \\ contradict the source text\end{tabular}             \\ \bottomrule
\end{tabular}%
}
\caption{Faithfulness error categories.}
\label{tab:error-categories}
\end{table}

\subsection{Key-fact Alignment}
The key-fact alignment task is designed to compute recall and multi-level faithfulness scores. Table~\ref{tab:key_fact_alignment_eval} provides the key-fact alignment prompt. The corresponding key-fact tree is linearized into a key-fact list using depth-first traversal. Along with the original summary sentences, this list is provided to GPT-4o as the evaluator to determine whether each key-fact is included (1) or not included (0) in the summary. Additionally, the sentence number from the summary that contains each key-fact is recorded and aligned with the corresponding key-fact. This can be further used for multi-level recall and multi-level faithfulness scores calculation.

\begin{table*}[ht!]
\centering
\tiny
\setlength{\tabcolsep}{1pt}%
\begin{tabular}{@{}llcccccccccccccccccccccccc@{}}
\toprule
\multicolumn{1}{c}{\multirow{2}{*}{Perspective}} &
  \multicolumn{1}{c}{\multirow{2}{*}{Model}} &
  \multicolumn{4}{c}{Position 0-20\%} &
  \multicolumn{4}{c}{Position 20-40\%} &
  \multicolumn{4}{c}{Position 40-60\%} &
  \multicolumn{4}{c}{Position 60-80\%} &
  \multicolumn{4}{c}{Position 80-100\%} &
  \multicolumn{4}{c}{Average} \\ \cmidrule(l){3-26} 
\multicolumn{1}{c}{} &
  \multicolumn{1}{c}{} &
  \multicolumn{1}{l}{Root} &
  \multicolumn{1}{l}{Branch} &
  \multicolumn{1}{l}{Leaf} &
  \multicolumn{1}{l}{Whole} &
  \multicolumn{1}{l}{Root} &
  \multicolumn{1}{l}{Branch} &
  \multicolumn{1}{l}{Leaf} &
  \multicolumn{1}{l}{Whole} &
  \multicolumn{1}{l}{Root} &
  \multicolumn{1}{l}{Branch} &
  \multicolumn{1}{l}{Leaf} &
  \multicolumn{1}{l}{Whole} &
  \multicolumn{1}{l}{Root} &
  \multicolumn{1}{l}{Branch} &
  \multicolumn{1}{l}{Leaf} &
  \multicolumn{1}{l}{Whole} &
  \multicolumn{1}{l}{Root} &
  \multicolumn{1}{l}{Branch} &
  \multicolumn{1}{l}{Leaf} &
  \multicolumn{1}{l}{Whole} &
  \multicolumn{1}{l}{Root} &
  \multicolumn{1}{l}{Branch} &
  \multicolumn{1}{l}{Leaf} &
  \multicolumn{1}{l}{Whole} \\ \midrule
\multirow{7}{*}{Analytical} &
  GTP-4o &
  0.87 &
  0.53 &
  0.18 &
  0.35 &
  0.85 &
  0.50 &
  0.16 &
  0.33 &
  0.86 &
  0.53 &
  0.17 &
  0.34 &
  0.88 &
  0.49 &
  0.18 &
  0.35 &
  0.89 &
  0.53 &
  0.20 &
  0.37 &
  0.87 &
  0.52 &
  0.18 &
  0.35 \\
 &
  GPT-4o-Mini &
  0.87 &
  0.53 &
  0.17 &
  0.35 &
  0.87 &
  0.50 &
  0.14 &
  0.33 &
  0.85 &
  0.51 &
  0.16 &
  0.33 &
  0.90 &
  0.51 &
  0.16 &
  0.34 &
  0.84 &
  0.47 &
  0.16 &
  0.32 &
  0.86 &
  0.50 &
  0.16 &
  0.33 \\
 &
  Claude-3.5-Sonnet &
  0.82 &
  0.53 &
  0.19 &
  0.36 &
  0.84 &
  0.53 &
  0.17 &
  0.35 &
  0.80 &
  0.52 &
  0.17 &
  0.33 &
  0.82 &
  0.51 &
  0.19 &
  0.35 &
  0.81 &
  0.54 &
  0.27 &
  0.41 &
  0.82 &
  0.53 &
  0.20 &
  0.36 \\
 &
  Claude-3.5-Haiku &
  0.80 &
  0.50 &
  0.17 &
  0.34 &
  0.80 &
  0.47 &
  0.14 &
  0.31 &
  0.76 &
  0.50 &
  0.17 &
  0.32 &
  0.80 &
  0.45 &
  0.16 &
  0.31 &
  0.78 &
  0.53 &
  0.23 &
  0.38 &
  0.79 &
  0.49 &
  0.18 &
  0.33 \\
 &
  LLaMA-3.1-405b-Inst. &
  0.88 &
  0.51 &
  0.15 &
  0.33 &
  0.88 &
  0.48 &
  0.14 &
  0.32 &
  0.84 &
  0.51 &
  0.15 &
  0.32 &
  0.86 &
  0.45 &
  0.15 &
  0.31 &
  0.87 &
  0.51 &
  0.18 &
  0.35 &
  0.87 &
  0.49 &
  0.15 &
  0.33 \\
 &
  LLaMA-3.1-8b-Inst. &
  0.88 &
  0.47 &
  0.12 &
  0.31 &
  0.86 &
  0.43 &
  0.11 &
  0.29 &
  0.84 &
  0.49 &
  0.12 &
  0.29 &
  0.90 &
  0.47 &
  0.15 &
  0.33 &
  0.85 &
  0.47 &
  0.15 &
  0.32 &
  0.87 &
  0.47 &
  0.13 &
  0.31 \\ \cmidrule(l){2-26} 
 &
  Average &
  0.85 &
  0.51 &
  0.16 &
  0.34 &
  0.85 &
  0.49 &
  0.14 &
  0.32 &
  0.83 &
  0.51 &
  0.15 &
  0.32 &
  0.86 &
  0.48 &
  0.17 &
  0.33 &
  0.84 &
  0.51 &
  0.20 &
  0.36 &
  0.85 &
  0.50 &
  0.17 &
  0.33 \\ \midrule
\multirow{7}{*}{Narrative} &
  GTP-4o &
  0.91 &
  0.58 &
  0.29 &
  0.45 &
  0.86 &
  0.56 &
  0.27 &
  0.43 &
  0.93 &
  0.54 &
  0.27 &
  0.43 &
  0.89 &
  0.51 &
  0.26 &
  0.42 &
  0.93 &
  0.64 &
  0.34 &
  0.50 &
  0.90 &
  0.56 &
  0.29 &
  0.45 \\
 &
  GPT-4o-Mini &
  0.89 &
  0.51 &
  0.23 &
  0.40 &
  0.85 &
  0.47 &
  0.21 &
  0.36 &
  0.90 &
  0.45 &
  0.19 &
  0.36 &
  0.87 &
  0.46 &
  0.20 &
  0.37 &
  0.86 &
  0.52 &
  0.21 &
  0.39 &
  0.88 &
  0.48 &
  0.21 &
  0.38 \\
 &
  Claude-3.5-Sonnet &
  0.84 &
  0.57 &
  0.28 &
  0.44 &
  0.84 &
  0.54 &
  0.28 &
  0.43 &
  0.90 &
  0.55 &
  0.27 &
  0.44 &
  0.84 &
  0.51 &
  0.25 &
  0.41 &
  0.89 &
  0.64 &
  0.38 &
  0.53 &
  0.86 &
  0.56 &
  0.30 &
  0.45 \\
 &
  Claude-3.5-Haiku &
  0.79 &
  0.51 &
  0.26 &
  0.39 &
  0.77 &
  0.52 &
  0.23 &
  0.38 &
  0.87 &
  0.49 &
  0.26 &
  0.40 &
  0.79 &
  0.47 &
  0.21 &
  0.37 &
  0.86 &
  0.64 &
  0.34 &
  0.50 &
  0.82 &
  0.53 &
  0.26 &
  0.41 \\
 &
  LLaMA-3.1-405b-Inst. &
  0.88 &
  0.55 &
  0.23 &
  0.40 &
  0.87 &
  0.52 &
  0.20 &
  0.38 &
  0.95 &
  0.51 &
  0.23 &
  0.41 &
  0.89 &
  0.52 &
  0.25 &
  0.41 &
  0.90 &
  0.62 &
  0.32 &
  0.48 &
  0.90 &
  0.54 &
  0.24 &
  0.42 \\
 &
  LLaMA-3.1-8b-Inst. &
  0.86 &
  0.43 &
  0.15 &
  0.32 &
  0.80 &
  0.43 &
  0.16 &
  0.32 &
  0.91 &
  0.44 &
  0.17 &
  0.34 &
  0.83 &
  0.43 &
  0.20 &
  0.35 &
  0.90 &
  0.56 &
  0.27 &
  0.44 &
  0.86 &
  0.46 &
  0.19 &
  0.36 \\ \cmidrule(l){2-26} 
 &
  Average &
  0.86 &
  0.53 &
  0.24 &
  0.40 &
  0.83 &
  0.51 &
  0.22 &
  0.38 &
  0.91 &
  0.50 &
  0.23 &
  0.40 &
  0.85 &
  0.48 &
  0.23 &
  0.39 &
  0.89 &
  0.60 &
  0.31 &
  0.47 &
  0.87 &
  0.52 &
  0.25 &
  0.41 \\ \bottomrule
\end{tabular}%
\caption{Model-wise breakdown of recall across key-fact levels, relative positions of information in input context, and summary perspectives.}
\label{tab:recall_total}
\end{table*}

\section{Annotator Recruitment Details}\label{apppendix:annotator_recruitment}

\subsection{Expert Annotators}
We recruited three postgraduate students specializing in Natural Language Processing as expert annotators (with C2-level English proficiency) for the validation of the three critical components of our automated pipeline: key-fact tree generation, query generation, and automated evaluation (see Section~\ref{sec:expert_eval}). They were compensated at a rate of \$30 per hour, with additional performance-based incentives to ensure high-quality contributions.

\subsection{Crowd-sourced Annotators}\label{appendix:crowd_sourced_anno}
As a baseline for our automated summarization with GPT-4o, we used summary evaluations from crowd-sourced annotators (see Section~\ref{sec:expert_eval}). For this purpose, we recruited three annotators for every Human Intelligence Task (HIT) from Amazon Mechanical Turk (MTurk) who met stringent qualification requirements.

Our recruitment criteria included successful completion of an English comprehension assessment that mirrored the actual annotation tasks of fact verification and key-fact alignment. Additionally, workers were required to maintain a minimum 90\% lifetime approval rate and demonstrate experience with at least 500 previously accepted HITs. All annotators received compensation exceeding the U.S. federal minimum wage.

For quality control, we embedded 5--10\% hidden attention-check questions with predetermined answers within each HIT. Any submissions failing these attention checks were rejected. This rigorous quality assurance procedure effectively filtered unreliable responses and ensured that all collected annotations were of high quality.

\section{Supplementary Result}\label{appendix:supplement}

We provide supplementary result that provides further insights into our primary findings. 

\begin{table}[ht]
\centering
\footnotesize
\begin{tabular}{@{}lc@{}}
\toprule
Category                                                                 & \% of Total Faithfulness Errors \\ \midrule
\begin{tabular}[c]{@{}l@{}}(A) Extrinsic Error \end{tabular}                                                                 & 95\%                                      \\
(B) Intrinsic Error                                                                  & 5\%                                     \\ \midrule
\begin{tabular}[c]{@{}l@{}}Subcategories of \\ (B) Intrinsic Error\end{tabular} & \% of (B) Intrinsic Error       \\ \midrule
Relation Error                                                                         & 54\%                                     \\
Entity Error                                                                       & 45\%                                     \\
Sentence Error                                                                       & 1\%                                      \\ \bottomrule
\end{tabular}%
\caption{Faithfulness error type distribution. The result is averaged over the six summarizers, excluding the two Qwen-2.5-32B models (Instruct and R1-distill).}
\label{tab:error_type}
\end{table}

\subsection{Evaluation Performance}\label{appendix:evaluation_performance}

Table~\ref{tab:recall_total} shows the detailed model-wise breakdown of key-fact retention performance (\emph{i.e.}, multi-level recall) across key-fact levels and relative positions of information in input context, and summary perspectives

\subsection{Faithfulness Error Distribution}\label{appendix:faithfulness_error_analysis}

Table \ref{tab:error_type} reveals that extrinsic errors constitute 95\% of all faithfulness errors across the evaluated summarization systems, with intrinsic errors accounting for only 5\%. Within the category of intrinsic errors, relation errors (54\%) and entity errors (45\%) account for nearly all cases, with sentence errors representing only 1\%.

The overwhelming prevalence of extrinsic errors suggests that current LLMs have a fundamental tendency to generate plausible but unfounded content, rather than merely misrepresenting information that exists in the source. Notably, the near absence of sentence errors (only 1\% of intrinsic errors) indicates that even when summarizing book-length content, models rarely produce statements that completely contradict the source material. Instead, when errors do occur within the bounds of input context, they typically manifest as more nuanced misrepresentations of specific entities or their relationships, rather than wholesale errors.

\clearpage
\begin{table*}[htbp]
\centering
\begin{tabular}{p{0.95\textwidth}}
\toprule

\hspace{0.2cm}You will be given an excerpt of a longer text. Read the excerpt carefully and extract all the key analytical insights related to the structure, relationships, and significance of the content. Organize these insights into a hierarchical tree structure with three levels: Root, Branches, and Leaves. \\

\vspace{0.1cm}
Structure levels:
\\
\vspace{-0.6cm}
\begin{itemize}
    \item[$\bullet$] Root: A single concise sentence summarizing the overarching purpose, argument, or main analytical insight of the text.
    \item[$\bullet$] Branches: Key supporting ideas, arguments, or elements that develop the overarching purpose or insight, including significant stages, relationships, or turning points in the text.
    \item[$\bullet$] Leaves: Specific evidence, minor details, or examples that provide additional support or elaboration for each branch.
\end{itemize}
\\

Requirements:
\\
\vspace{-0.6cm}
\begin{enumerate}
    \item Do not omit any significant information from the text.
    \item Ensure clear relationships between roots, branches, and leaves.
    \item All key-facts must be directly supported by the text.
    \item Create as many roots, branches and leaves as needed to fully capture the text's key-facts.
    \item All key-facts should NEVER be based on over-interpretation or logical leaps beyond the information provided in the text.
    \item Focus on analyzing what is explicitly stated, supported, or implied within reasonable bounds, without adding subjective opinions or unsupported inferences.
    \item NEVER use pronouns, such as he, she, it, that, or "the protagonist". ALWAYS USE PROPER NOUNS.
    \item Make each key-fact as concise as possible, ensuring that each contain at most 2-3 entities.
\end{enumerate}

\vspace{0.1cm}

Output format: 
\\
\hspace{0.2cm} - Provide your answer in JSON format. \\
\hspace{0.2cm} - The answer should ONLY be a dictionary with the valid JSON format as follows: <Tree> \\
\hspace{0.2cm} - Include only the tree dictionary in the answer. \\

\vspace{0.1cm}
The excerpt: \\
\hspace{0.2cm} \textcolor{blue}{$\{$excerpt$\}$} \\
\bottomrule
\end{tabular}
\caption{Analytical key-fact tree generation prompt.}
\label{tab:analytical_tree_generation}
\end{table*}

\begin{table*}[htbp]
\centering
\begin{tabular}{p{0.95\textwidth}}
\toprule
\hspace{0.2cm}You will be given an excerpt of a longer text. Read the excerpt carefully and extract all the key-facts related to the sequence of events and key developments in a straightforward and chronological manner. Organize these key-facts into a hierarchical tree structure with three levels: Root, Branches, and Leaves. \\
\vspace{0.1cm}

Structure levels: \\
\vspace{-0.6cm}
\begin{itemize}
    \item[$\bullet$] Root: A single concise sentence capturing the main idea or overarching sequence of events in the text.
    \item[$\bullet$] Branches: Key supporting events or developments that progress the narrative logically, including major stages, actions or transitions.
    \item[$\bullet$] Leaves: Specific details, minor events, or pieces of evidence that provide additional clarity or elaboration for each branch.
\end{itemize}
\\

Requirements:
\\
\vspace{-0.6cm}

\begin{enumerate}
    \item Each key-fact is NOT a statement of theme or topic of the text, but a specific piece of information that can be directly extracted from the text.
    \item Do not omit any significant information from the text.
    \item Ensure clear relationships between roots, branches, and leaves.
    \item All key-facts must be directly supported by the text.
    \item Create as many roots, branches and leaves as needed to fully capture the text's key-facts.
    \item Focus on how the story progresses from beginning to end, including any critical pivots or climaxes.
    \item NEVER use pronouns, such as he, she, it, that, or "the protagonist". ALWAYS USE PROPER NOUNS.
    \item Make each key-fact as concise as possible, ensuring that each contain at most 2-3 entities.
\end{enumerate}

Output format: \\
\hspace{0.2cm} - Provide your answer in JSON format. \\
\hspace{0.2cm} - The answer should ONLY be a dictionary with the valid JSON format as follows: <Tree> \\
\hspace{0.2cm} - Include only the tree dictionary in the answer. \\

\vspace{0.1cm}
The excerpt: \\
\hspace{0.2cm}  \textcolor{blue}{$\{$excerpt$\}$} \\
\bottomrule
\end{tabular}
\caption{Narrative key-fact tree generation prompt.}
\label{tab:narrative_tree_generation}
\end{table*}
\begin{table*}[htbp]
\centering
\begin{tabular}{p{0.95\textwidth}}
\toprule
You will receive: 
\\
\vspace{-0.6cm}
\begin{itemize}
    \item[--] An excerpt from a novel (the source text). 
    \item[--] A hierarchical key-fact tree in JSON format, structured as follows:  <Tree>
\end{itemize}

\vspace{0.1cm}
Your task:
\\
\vspace{-0.6cm}
\begin{itemize}
    \item[$\bullet$] Evaluate the faithfulness of each fact in the key-fact tree by carefully comparing it to the provided novel excerpt. 
    \item[$\bullet$] For every root key-fact, branch key-fact, and leaf key-fact, assign a binary label based on the following criteria:
    \begin{itemize}
        \item[--] 1 (Faithful): The fact is fully accurate and directly supported by the text.
        \item[--] 0 (Unfaithful): The fact is inaccurate, misleading, or not supported by the text.
    \end{itemize}
    \item[$\bullet$] For each evaluation, provide a justification explaining why the fact is marked as Faithful or Unfaithful. Condense your reason to one or two sentences.
\end{itemize}

\vspace{0.1cm}
Important guidelines: 
\\
\vspace{-0.6cm}
\begin{itemize}
    \item[$\bullet$] Only accept facts that are explicitly stated. Do not accept information that is inferred, altered, or expanded beyond what the text directly says.
    \item[$\bullet$] Be precise and consistent. Evaluate each level—roots, branches, and leaves—independently for accuracy.
    \item[$\bullet$] Maintain input structure: The output must preserve the exact same hierarchical structure as the input key-fact tree. Each node of the output tree should represent the label (0 or 1) and its corresponding justification for that key-fact.
\vspace{-0.2cm}
\end{itemize}

\vspace{0.1cm}
Output format: \\
\hspace{0.2cm} - Provide your answer in JSON format. \\
\hspace{0.2cm} - The answer should ONLY be a dictionary with the valid JSON format as follows: <Tree> \\
\hspace{0.2cm} - Include only the tree dictionary in the answer. \\

\vspace{0.1cm}
The excerpt: \\
\hspace{0.2cm}  \textcolor{blue}{$\{$excerpt$\}$} \\

\vspace{0.1cm}
The key-fact tree: \\
\hspace{0.2cm}  \textcolor{blue}{$\{$key-fact tree$\}$} \\
\bottomrule
\end{tabular}
\caption{Faithfulness evaluation of key-fact tree prompt.}
\label{tab:keyfact_tree_validation_faith}
\end{table*}
\begin{table*}[htbp]
\centering
\begin{tabular}{p{0.95\textwidth}}
\toprule
You will receive: 
\\
\vspace{-0.6cm}
\begin{itemize}
    \item[--] An excerpt from a novel (the source text). 
    \item[--] A hierarchical key-fact tree in JSON format, structured as follows: <Tree>
\end{itemize}

\vspace{0.1cm}
Your task: \\
\vspace{-0.6cm}
\begin{itemize}
    \item[$\bullet$] Evaluate the subjectivity of each fact in the key-fact tree by comparing it to the provided excerpt.
    \item[$\bullet$] For every root key-fact, branch key-fact, and leaf key-fact, assign a binary label based on the following criteria:
    \begin{itemize}
        \item[--] 1 (Objective): The statement is purely fact-based and directly supported by the text, without subjective language or interpretation.
        \item[--] 0 (Subjective): The statement includes opinions, assumptions, interpretations, or evaluative/adjectival language.
    \end{itemize}
    \item[$\bullet$] For each evaluation, provide a justification explaining why the fact is marked as objective or subjective. Condense your reason to one or two sentences.
\end{itemize}

\vspace{0.1cm}
Important guidelines: \\
\vspace{-0.6cm}
\begin{itemize}
    \item[$\bullet$] Accept only factual statements. Statements must reflect exactly what is stated in the text without additional interpretation.
    \item[$\bullet$] Reject subjective language. Statements that express opinions, emotions, or biases must be marked as 0.
    \item[$\bullet$] Evaluate each level independently. Assess the objectivity of roots, branches, and leaves separately.
    \item[$\bullet$] Maintain input structure: The output must preserve the exact same hierarchical structure as the input key-fact tree. Each node of the output should represent the label (0 or 1) and its corresponding justification for that key-fact.
\vspace{-0.2cm}
\end{itemize}

\vspace{0.1cm}
Output format: \\
\hspace{0.2cm} - Provide your answer in JSON format. \\
\hspace{0.2cm} - The answer should ONLY be a dictionary with the valid JSON format as follows: <Tree> \\
\hspace{0.2cm} - Include only the tree dictionary in the answer. \\

\vspace{0.1cm}
The excerpt: \\
\hspace{0.2cm}  \textcolor{blue}{$\{$excerpt$\}$} \\

\vspace{0.1cm}
The key-fact tree: \\
\hspace{0.2cm}  \textcolor{blue}{$\{$key-fact tree$\}$} \\
\bottomrule
\end{tabular}
\caption{Objectivity evaluation of key-fact tree prompt.}
\label{tab:keyfact_tree_validation_subj}
\end{table*}
\begin{table*}[htbp]
\centering
\begin{tabular}{p{0.95\textwidth}}
\toprule
You will receive: 
\\
\vspace{-0.6cm}
\begin{itemize}
    \item[--] An excerpt from a novel (the source text). 
    \item[--] A hierarchical key-fact tree in JSON format, structured as follows:  <Tree>
\end{itemize}
\vspace{-0.2cm}

\vspace{0.1cm}
Your task: \\
\vspace{-0.6cm}
\begin{itemize}
    \item[$\bullet$] Evaluate the significance of each fact in the key-fact tree based on the provided excerpt.
    \item[$\bullet$] For every root key-fact, branch key-fact, and leaf key-fact, assign a binary label based on the following criteria:
    \begin{itemize}
        \item[--] 1 (Significant): The fact is essential for understanding the text, such as driving the plot forward, developing characters, or revealing major conflicts.
        \item[--] 0 (Insignificant): The fact is trivial, background information, or does not meaningfully contribute to the story's progression or understanding.
    \end{itemize}
    \item[$\bullet$] For each evaluation, provide a justification explaining why the fact is marked as significant or insignificant. Condense your reason to one or two sentences.
\end{itemize}

\vspace{0.1cm}
Important guidelines: \\
\vspace{-0.6cm}
\begin{itemize}
    \item[$\bullet$] Avoid trivial details: Facts that describe minor settings, insignificant actions, or irrelevant background information should be scored 0.
    \item[$\bullet$] Evaluate independently: Assess the significance of each root, branch, and leaf on its own merit.
    \item[$\bullet$] Maintain input structure: The output must preserve the exact same hierarchical structure as the input key-fact tree. Each node of the output should represent the label (0 or 1) and its corresponding justification for that key-fact.
\vspace{-0.2cm}
\end{itemize}

\vspace{0.1cm}
Output format: \\
\hspace{0.2cm} - Provide your answer in JSON format. \\
\hspace{0.2cm} - The answer should ONLY be a dictionary with the valid JSON format as follows: <Tree> \\
\hspace{0.2cm} - Include only the tree dictionary in the answer \\

\vspace{0.1cm}
The excerpt: \\
\hspace{0.2cm}  \textcolor{blue}{$\{$excerpt$\}$} \\

\vspace{0.1cm}
The key-fact tree: \\
\hspace{0.2cm}  \textcolor{blue}{$\{$key-fact tree$\}$} \\
\bottomrule
\end{tabular}
\caption{Significance evaluation of key-fact tree prompt.}
\label{tab:keyfact_tree_validation_sig}
\end{table*}
\begin{table*}[htbp]
\centering
\begin{tabular}{p{0.95\textwidth}}
\toprule
Main objective: \\
    \hspace{0.2cm} Craft a single query that requests a summary of the analytical content represented by the key-fact tree. The query should address the overarching purpose or argument (Root), the supporting ideas or elements (Branches), and the specific evidence or examples (Leaves), guiding a coherent examination of how each component relates to the text's main insight.\\

\vspace{0.1cm}
Definition of a key-fact tree: \\
\hspace{0.2cm} A key-fact tree is a hierarchical representation of the important information in a text, organized into three levels:
\vspace{-0.2cm}
\begin{itemize}
    \item[$\bullet$] Root: A single concise sentence summarizing the overarching purpose, argument, or main analytical insight of the text.
    \item[$\bullet$] Branches: Key supporting ideas, arguments, or elements that develop the overarching purpose or insight, including significant stages, relationships, or turning points in the text.
    \item[$\bullet$] Leaves: Specific evidence, details, or examples that provide additional support or elaboration for each branch.
\end{itemize}

\vspace{0.1cm}
You will receive: \\
\vspace{-0.4cm}
\begin{itemize}
    \item[--] An excerpt from a text (the source text).
    \item[--] A tree of key-facts in JSON format, with the structure: <Tree>
\end{itemize}

\vspace{0.1cm}
Requirements: \\
\vspace{-0.4cm}
\begin{itemize}
    \item[--] The query should naturally lead to an answer that integrates the key-facts, showing how each piece of evidence or argument reinforces the main insight.
    \item[--] Your query should be specific enough to address the contents in the key-fact tree.
    \item[--] Make ONLY ONE query for the entire tree.
    \item[--] Your query should be as concise as possible.
    \item[--] Your query should NEVER mention anything about the key-fact tree.
\end{itemize}
\vspace{-0.2cm}

\vspace{0.1cm}
The excerpt: \\
\hspace{0.2cm}  \textcolor{blue}{$\{$excerpt$\}$} \\

\vspace{0.1cm}
The key-fact Tree: \\
\hspace{0.2cm} \textcolor{blue}{$\{$key-fact tree$\}$} \\
\bottomrule
\end{tabular}
\caption{Analytical query generation prompt.}
\label{tab:analytical_query_generation}
\end{table*}
\begin{table*}[htbp]
\centering
\begin{tabular}{p{0.95\textwidth}}
\toprule
Main objective: \\
    \hspace{0.2cm} Craft a single query that requests a summary of the narrative content represented by the keyfact tree. The query should address the overarching sequence of events (Root), the key supporting developments (Branches), and the specific details or events (Leaves), guiding a comprehensive and chronological explanation of how each component contributes to the overall narrative.\\

\vspace{0.1cm}
Definition of a key-fact tree: \\
\hspace{0.2cm} A key-fact tree is a hierarchical representation of the important information in a text, organized into three levels:
\vspace{-0.2cm}
\begin{itemize}
    \item[$\bullet$] Root: A single concise sentence capturing the main idea or overarching sequence of events in the text
    \item[$\bullet$] Branches: Key supporting events or developments that progress the narrative logically, including major stages, actions or transitions 
    \item[$\bullet$] Leaves: Specific details, events, or pieces of evidence that provide additional clarity or elaboration for each branch
\end{itemize}

\vspace{0.1cm}
You will receive: \\
\vspace{-0.4cm}
\begin{itemize}
    \item[--] An excerpt from a text (the source text).
    \item[--] A tree of key-facts in JSON format, with the structure: <Tree>
\end{itemize}

\vspace{0.1cm}
Requirements: \\
\vspace{-0.4cm}
\begin{itemize}
    \item[--] The query should naturally lead to an answer that integrates the key-facts in a coherent, chronological narrative.
    \item[--] Your query should be specific enough to address the contents in the key-fact tree.
    \item[--] Make ONLY ONE query for the entire tree.
    \item[--] Your query should be as concise as possible.
    \item[--] Your query should NEVER mention anything about the key-fact tree.
\end{itemize}

\vspace{0.1cm}
The excerpt: \\
\hspace{0.2cm}  \textcolor{blue}{$\{$excerpt$\}$} \\

\vspace{0.1cm}
The key-fact tree: \\
\hspace{0.2cm} \textcolor{blue}{$\{$key-fact tree$\}$} \\
\bottomrule
\end{tabular}
\caption{Narrative query generation prompt.}
\label{tab:narrative_query_generation}
\end{table*}
\begin{table*}[htbp]
\centering
\begin{tabular}{p{0.95\textwidth}}
\toprule
You will receive an excerpt of a novel and its corresponding summary, split into multiple sentences. Your task is to assess how faithfully each summary sentence represents the given excerpt's content.\\

Faithfulness means the summary accurately reflects the information and meaning conveyed in the excerpt, without introducing unsupported claims or contradicting the source material.\\
\vspace{-0.2cm}
When evaluating faithfulness, your decision should be one of these error categories: \\
\vspace{-0.4cm}
\begin{itemize}
    \item[$\bullet$] Out-of-article error: The summary introduces facts, opinions, or information not found in or reasonably inferrable from the text.
    \item[$\bullet$] Entity error: Incorrect reference to key subjects/objects (e.g., wrong names, numbers, pronouns).
    \item[$\bullet$] Relation error: Mistakes in semantic relationships (e.g., incorrect verbs, prepositions, adjectives).
    \item[$\bullet$] Sentence error: Multiple errors causing the entire sentence to contradict the text.
    \item[$\bullet$] No error: The summary statement aligns with the text's content.
\end{itemize}
\vspace{-0.2cm}
Guidelines for evaluating abstractive summaries: \\
\vspace{-0.4cm}
\begin{itemize}
    \item[$\bullet$] Logical inference: If the summary makes reasonable conclusions based on information presented in the text, mark it as faithful (no error).
    \item[$\bullet$] Paraphrasing: Different word choices or sentence structures that preserve the original meaning are faithful.
    \item[$\bullet$] Generalization: Combining multiple specific details into a broader statement is faithful if accurate.
    \item[$\bullet$] Implicit information: Drawing on clearly implied information from context is faithful.
\end{itemize}

\vspace{0.1cm}
Instruction: \\
\vspace{-0.4cm}
\begin{itemize}
    \item[--] Compare each summary sentence with the text.
    \item[--] Provide a single, concise sentence explaining any factuality error, referencing specific elements from both texts.
    \item[--] Classify the error category for each sentence.
\end{itemize}

\vspace{0.1cm}
Please provide your answer in JSON format as a list of dictionaries with keys "sentence", "reason", and "category" as follows: 

\\

[\{\\
"sentence": "first sentence",\\
"reason": "your reason",\\
"category": "error category"\\
\}, \{\\
"sentence": "second sentence",\\
"reason": "your reason",\\
"category": "error category"\\
\}]\\

\vspace{0.1cm}
Excerpt:\\

\hspace{0.2cm}  \textcolor{blue}{$\{$excerpt$\}$} \\
Summary with \textcolor{blue}{$\{$$\#$ sentences$\}$} sentences:\\
\hspace{0.2cm} \textcolor{blue}{$\{$summary sentences$\}$}\\

\bottomrule

\end{tabular}
\caption{Summary evaluation prompt: fact-verification.}
\label{tab:fact_checking_eval}
\end{table*}
\begin{table*}[htbp]
\centering
\begin{tabular}{p{0.95\textwidth}}
\toprule
\hspace{0.2cm} You will receive a summary and a set of key-facts for the same transcript. Your task is to assess if each key-fact is inferred from the summary.\\
\vspace{-0.2cm}
Instruction: \\
\vspace{-0.4cm}
\begin{itemize}
    \item[--] First, compare each key-fact with the summary.
    \item[--] Second, check if the key-fact is inferred from the summary and then respond "Yes" or "No" for each key-fact. If "Yes", specify the line number(s) of the summary sentence(s) relevant to each key-fact. 
\end{itemize}
\vspace{-0.2cm}
Provide your answer in JSON format. The answer should be a list of dictionaries whose keys are "key-fact", "response", and "line number":\\
\vspace{0.2cm}

[\{"key-fact": "first key-fact", "response": "Yes", "line number": [1]\}, \\
\{"key-fact": "second key-fact", "response": "No", "line number": []\}, \\
\{"key-fact": "third key-fact", "response": "Yes", "line number": [1, 2, 3]\}]\\

\vspace{0.2cm}
Summary:\\
\hspace{0.2cm}  \textcolor{blue}{$\{$summary$\}$} \\

\vspace{0.2cm}

 \textcolor{blue}{$\{$\# key-facts$\}$} key-facts:\\
\hspace{0.2cm}  \textcolor{blue}{$\{$key-fact list$\}$} \\
\bottomrule
\end{tabular}
\caption{Summary evaluation prompt: key-fact alignment.}
\label{tab:key_fact_alignment_eval}
\end{table*}
\begin{table*}[htbp]
\centering
\begin{tabular}{p{0.95\textwidth}}
\toprule
You are an expert evaluator of novel chunks. 
Read the entire chunk, perform all reasoning silently, and output only the final JSON object described below—no other text.\\
DEFINITIONS \\
\begin{itemize}
    \item[$\bullet$] Major fact (MF) = a distinct event, statement, or data point that introduces new information (e.g., “John confesses the theft,” “The storm destroys the lighthouse”).  Repeated or trivial details do not count.
    \item[$\bullet$] Internal connection (IC) = an explicit or implicit link between two MFs that shows foreshadowing, cause-effect, contrast, or resolution across sentences or paragraphs.
\end{itemize}
SCORING RUBRICS\\
\begin{enumerate}
    \item HKF\_validity ― Validity of Hierarchical Key-Fact Tree Extraction Count the MFs and judge whether they can be arranged into a clear multi-level tree (root → branches → leaves).
        \begin{itemize}
            \item Less than 25 MFs  →  Score 1-2 
            \item 25 - 35 MFs and at least a two-level hierarchy → Score 3-4 
            \item More than 35 MFs and a well-defined multi-level hierarchy → Score 5 
        \end{itemize}
    \item Content\_coherence - Evaluate the chunk using the following four signals: 
    \begin{itemize}
        \item A. Structural completeness (beginning → middle → end)      : YES / NO 
        \item B. Abrupt transitions (N\_abrupt)                           : 0, 1-2, ≥3
        \item C. Unresolved references or unfinished plotlines           : 0, 1, ≥2
        \item D. Logical/temporal/spatial incoherences (N\_incoherent)    : 0, 1, ≥2
    \end{itemize}
   
   Scoring
        \begin{itemize}
            \item 1 point= A = NO and (N\_abrupt ≥3 or N\_unresolved ≥2 or N\_incoherent ≥2)
            \item 2 points= A = NO and at least two of B-D are in the "1-2 / 1" range
            \item 3 points= A = YES but exactly one of B-D in the "1-2 / 1" range
            \item 4 points= A = YES and at most one mild issue (B-D = 1-2 / 1); others 0
            \item 5 points= A = YES and N\_abrupt = N\_unresolved = N\_incoherent = 0
        \end{itemize}
    \item Cross\_content\_reasoning ― Support for Cross-Content Reasoning
   Count the ICs that enable reasoning across different parts of the chunk.
        \begin{itemize}
            \item Less than 8 ICs → Score 1-2  
            \item 8 - 16 ICs → Score 3-4  
            \item More than 16 ICs  → Score 5  
        \end{itemize}
\end{enumerate}

\\
OUTPUT FORMAT: \\ 
\\

\{\\
  "HKF\_validity": <score from 1 to 5>,\\
  "Content\_coherence": <score from 1 to 5>,\\
  "Cross\_content\_reasoning": <score from 1 to 5>,\\
  "explanation": "Less than 25-word justification"\\
\}\\

INPUT:\\
 \textcolor{blue}{$\{$chunk$\}$}\\

\bottomrule

\end{tabular}
\caption{Chunk size validation prompt.}
\label{tab:chunk_size_prompt}
\end{table*}

\end{document}